\documentclass{article}
\usepackage[T1]{fontenc}
\usepackage[preprint]{iclr2027_conference}
\usepackage{times}

\usepackage{hyperref}

\usepackage{amsmath,amsfonts,bm}

\def\eqref#1{equation~\ref{#1}}

\def\1{\bm{1}}

\DeclareMathAlphabet{\mathsfit}{\encodingdefault}{\sfdefault}{m}{sl}
\SetMathAlphabet{\mathsfit}{bold}{\encodingdefault}{\sfdefault}{bx}{n}

\usepackage{multirow}
\usepackage{graphicx}
\usepackage{subfigure}
\usepackage{amsmath}
\usepackage{amssymb}
\usepackage{mathtools}
\usepackage{amsthm}
\usepackage[capitalize,noabbrev]{cleveref}
\usepackage{fancyhdr}
\usepackage{algorithm}
\usepackage{eso-pic}
\usepackage{forloop}
\usepackage{url}
\usepackage{makecell}

\usepackage[normalem]{ulem}
\usepackage{stfloats}

\theoremstyle{plain}

\theoremstyle{definition}

\theoremstyle{remark}

\usepackage{algpseudocode}
\usepackage{caption}

\algrenewcommand\algorithmicrequire{\textbf{Input:}}
\algrenewcommand\algorithmicensure{\textbf{Output:}}

\usepackage[utf8]{inputenc}
\usepackage{newunicodechar}
\newunicodechar{−}{\textminus}

\usepackage{url}
\usepackage{booktabs}
\usepackage{tabularx}
\usepackage{longtable}
\DeclareMathSizes{7.5}{7}{5}{5}
\newcommand{\name}{HyperRepair}
\newcommand{\repairman}{RSN}
\newcommand{\hypernet}{HyperNet}
\newcommand{\papertitle}{\name{}: Recovering Forgotten General Capabilities of Fine-Tuned Language Models}

\title{\papertitle}

\author{%
\textbf{Junrong Tang$^{1,2}$, \quad Jianghui Wang$^{1}$, \quad Hao Kong$^{3}$, \quad Yang Lin$^{3}$, \quad Yibo Yang$^{1}$\thanks{Corresponding author.}}\\
$^{1}$Shanghai Jiao Tong University\\
$^{2}$The University of Queensland\\
$^{3}$Wanwu-MindFlow.ai\\
\texttt{yibo.yang@sjtu.edu.cn}
}

\begin{document}
\addtocontents{toc}{\protect\setcounter{tocdepth}{-1}}

\maketitle
\suppressfloats[t]

\begin{abstract}

Fine-tuning improves a language model's performance in a target domain but can weaken its general capabilities. Existing methods often face a trade-off between general capabilities and domain specialization, making it difficult to preserve both. We introduce HyperRepair, a framework for recovering general capabilities while retaining domain specialization at only inference cost. To make recovery generalizable across models sharing a pretrained backbone, HyperRepair learns a hypernetwork from repairs trained on source models. The hypernetwork uses activation fingerprints that summarize changes from the pretrained backbone to generate a tailored Repairing Side Net (\repairman{}), whose prompt-conditioned gate applies corrections selectively. This enables recovery on models from unseen adaptation domains without target-domain examples or additional optimization. We evaluate \name{} across two model families, 12 LoRA adaptation datasets, and 15 general benchmarks. On the default split, \name{} achieves 93.4\% of the base models' average general score while retaining 99.5\% of the fine-tuned models' average domain score.
\end{abstract}

\section{Introduction}
\label{sec:introduction}

The widespread use of pretrained large language models creates a need to adapt them to diverse downstream tasks, domains, and application requirements.
Fine-tuning improves downstream performance but can degrade pretrained knowledge and general capabilities \citep{wang2024twostage,kotha2024implicit,luo2025empirical}, a phenomenon known as catastrophic forgetting \citep{mccloskey1989catastrophic,yang2023neural,liu2025enhancing}.
Such degradation can impair responses beyond the adaptation task, limiting the usefulness of specialized models when applications also require broad knowledge and reasoning.
Although low-rank adaptation (LoRA) trains only a small number of additional parameters \citep{hu2022lora}, studies show that catastrophic forgetting can still occur \citep{kalajdzievski2024scaling,yang2024corda}.

Existing approaches mitigate forgetting during fine-tuning through parameter regularization \citep{kirkpatrick2017ewc}, rehearsal and distillation \citep{huang2024ssr,li2016lwf}, or changes to adapter construction and loss weighting \citep{yang2024corda,sanyal2025flow,lin2026talr}. However, these approaches intervene during adaptation and cannot directly repair an already fine-tuned model without further training. Post-hoc approaches instead modify trained weights through model merging \citep{ilharco2023taskarithmetic,yadav2023ties,huang2024emr,yu2024dare}, layerwise rescaling \citep{wang2025lines}, or forgetting-aware pruning \citep{huang2025fapm}. Post-hoc rescaling and pruning can recover general performance without repeating the original fine-tuning process. However, methods that apply a single set of modified weights to all inputs can face a trade-off between recovering general capabilities and retaining downstream specialization.
We therefore pose and study the following question:

\begin{quote}
\textit{Given a fine-tuned language model with degraded general capabilities, how can we efficiently recover its general performance at inference cost only while retaining downstream task performance?}
\end{quote}

\begin{figure}[t]
\centering
\includegraphics[width=\linewidth]{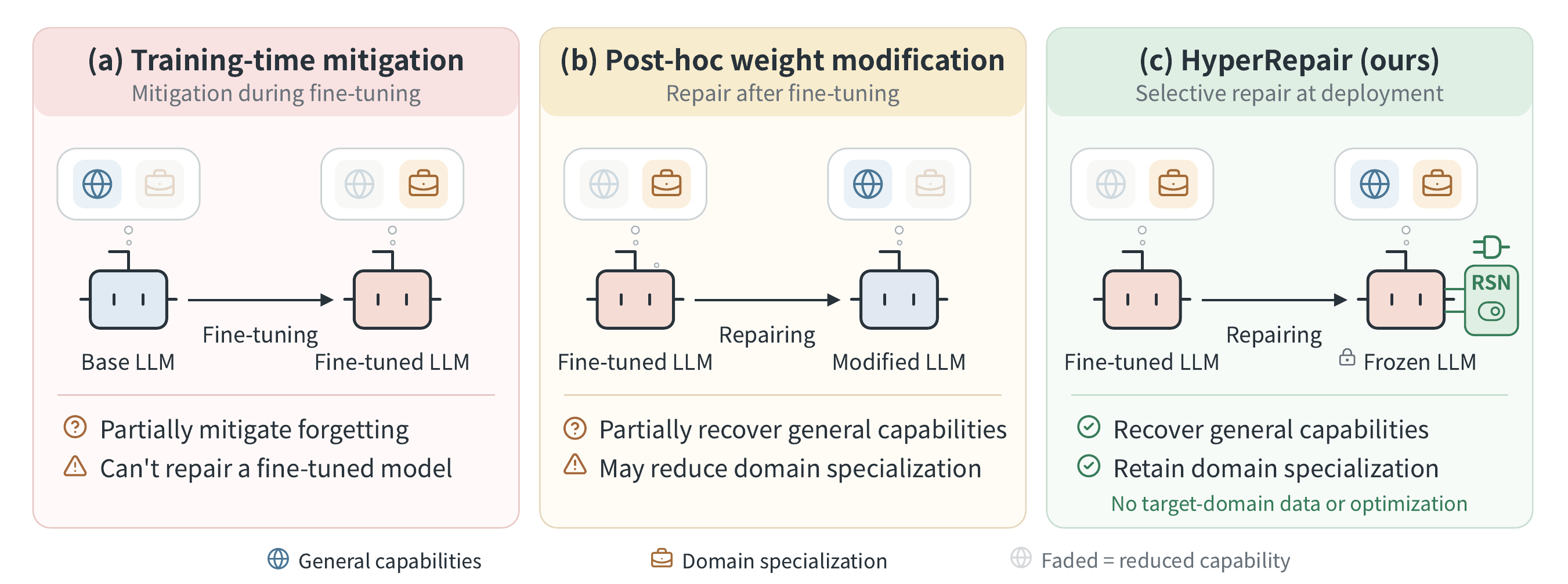}
\caption{Comparison of prior approaches and \name{}. (a) Training-time mitigation requires control over fine-tuning. (b) Post-hoc weight modification may trade domain specialization for general recovery. (c) \name{} selectively corrects a frozen model with a gated \repairman{} to recover general capabilities while retaining domain specialization, without target-domain data or optimization at deployment.}
\label{fig:intro}
\end{figure}

This motivates a selective approach to recovery: we keep the fine-tuned weights fixed and place the correction in a separate pathway (Figure~\ref{fig:intro}). This retains the original model as a fallback and makes it possible to decide from the prompt whether a correction should be used. Side networks provide a natural starting point for this design because they can reuse intermediate representations from a frozen backbone without rewriting its parameters \citep{zhang2020sidetuning,sung2022lst}. A separate pathway alone, however, does not solve the entire problem. If a new repair must be trained for every newly released adapter, the cost of recovery still grows with the number of models. What is missing is a way to learn from repaired models and use that experience to repair another model.

To make recovery reusable across models, we consider models adapted independently from the same pretrained backbone. Some of these models serve as sources from which repairs can be learned. The target model comes from an adaptation domain that was not used to learn the repair mechanism. Its domain data is unavailable, and no optimization is allowed on the target model. To our knowledge, this is the first work to study amortized recovery of general capabilities across independently fine-tuned language models. The goal is not merely to find a better compromise for one checkpoint, but to learn a repair process that generalizes to an unseen downstream domain.

Building on this formulation, we introduce \name{}, which learns this repair process. To prepare data for learning how to repair, we first train a lightweight Repairing Side Net (\repairman{}) for each fine-tuned source model. The \repairman{} reads intermediate hidden states and produces a residual correction before the language model head. Labeled general examples teach it how to improve general predictions. A prompt-conditioned gate determines how much of the correction should be applied. The gate is trained to activate on general prompts and to remain inactive on unlabeled domain prompts. When the correction is inactive, the model produces the original fine-tuned predictions. The model can therefore use the repair where it is helpful without giving up direct access to the behavior learned for a downstream domain.

The remaining question is how to obtain a suitable \repairman{} for a new model without training one. We represent each fine-tuned model by an activation fingerprint. The fingerprint records how its hidden representations differ from those of the pretrained model on a fixed group of unlabeled general prompts. We train a hypernetwork, \hypernet{} \citep{ha2017hypernetworks,aliev2026hypersafe}, to map this fingerprint to the complete parameters of the \repairman{}, including its prompt gate. Training uses the parameters of directly trained source repairs together with supervision on the behavior of the generated repairs. At deployment, \hypernet{} receives the fingerprint of a model from a held-out adaptation domain and generates its RSN in one forward pass. No target-domain examples or optimization are required.

We summarize our main contributions as follows:\par\nopagebreak[4]

\begin{enumerate}
\item We formulate post-hoc recovery as the activation-conditioned generation of model-specific Repairing Side Nets (\repairman{}s) that recover general capabilities while retaining downstream performance. Given a fine-tuned checkpoint, our method generates its repair from activation fingerprints without additional optimization or target-domain examples.

\item We design \name{}, a hypernetwork-based framework that predicts the weights of a lightweight \repairman{}, including its prompt-conditioned gate. The \repairman{} applies selective residual corrections to the frozen backbone, while \hypernet{} learns from source repairs and generalizes to new fine-tuned models.

\item We show that generated \repairman{}s can recover general capabilities while preserving domain performance. We evaluate two backbones across 12 adaptation datasets and 15 general benchmarks, training \hypernet{} on source models and testing on held-out domains. On the default split, \name{} reaches 93.4\% of the base models' average general score and retains 99.5\% of the fine-tuned models' average domain score.
\end{enumerate}

\section{Related Work}
\label{sec:related-work}

\paragraph{Forgetting in Fine-Tuned LLMs.}
Domain-adaptive pretraining \citep{gururangan2020dontstop} and supervised fine-tuning \citep{howard2018ulmfit} can improve target-task performance, while multitask instruction tuning \citep{wei2022flan,sanh2022t0} can improve generalization to unseen tasks. Fine-tuning can nevertheless degrade existing capabilities. Longer pretraining can increase sensitivity to parameter updates \citep{springer2025overtrained}, while smaller fine-tuning learning rates can reduce degradation \citep{lin2026talr}. Aggregate scores can conceal simultaneous gains and losses on individual examples \citep{harmon2026mapping}. Analyses of topic and style shifts \citep{zhang2024dissecting}, task inference \citep{kotha2024implicit}, function activation \citep{jiang2025functionvectors}, and task alignment \citep{zheng2025spurious} suggest that some declines reflect how retained knowledge is accessed and applied. We study whether general capabilities can be recovered after domain fine-tuning while retaining the domain capabilities acquired during adaptation.

\paragraph{Mitigating Fine-Tuning-Induced Forgetting.}
Classical continual learning mitigates forgetting through parameter regularization \citep{kirkpatrick2017ewc,zenke2017si,aljundi2018mas}, distillation \citep{li2016lwf}, gradient constraints \citep{lopezpaz2017gem,chaudhry2019agem}, and replay \citep{shin2017dgr,buzzega2020der}. LAMOL \citep{sun2020lamol} and Self-Synthesized Rehearsal \citep{huang2024ssr} generate examples for language-model rehearsal. CorDA's knowledge-preserved mode \citep{yang2024corda} initializes adapters using calibration activations, and FLOW \citep{sanyal2025flow} upweights examples with low pretrained-model loss. Post-hoc approaches include weight interpolation for visual robustness \mbox{\citep{wortsman2022wiseft}}, checkpoint averaging \citep{wortsman2022soups}, and merging with Fisher information \mbox{\citep{matena2022fisher}} or matching layer outputs \citep{jin2023regmean}. Several model-merging methods combine task updates \citep{ilharco2023taskarithmetic,yadav2023ties,yu2024dare,huang2024emr,yang2024adamerging}, while LiNeS \citep{wang2025lines} scales updates by layer depth. Side-Tuning \citep{zhang2020sidetuning} trains a task-specific side network and blends its features with those of a frozen pretrained model. Our RSNs recover general capabilities after fine-tuning and are generated by \hypernet{} for new models sharing the same backbone. \name{} generates a gated repair pathway alongside frozen fine-tuned weights, preserving the original predictions when correction is disabled.

\section{Method}
\label{sec:method}

\begin{figure}[t]
\centering
\includegraphics[width=\linewidth]{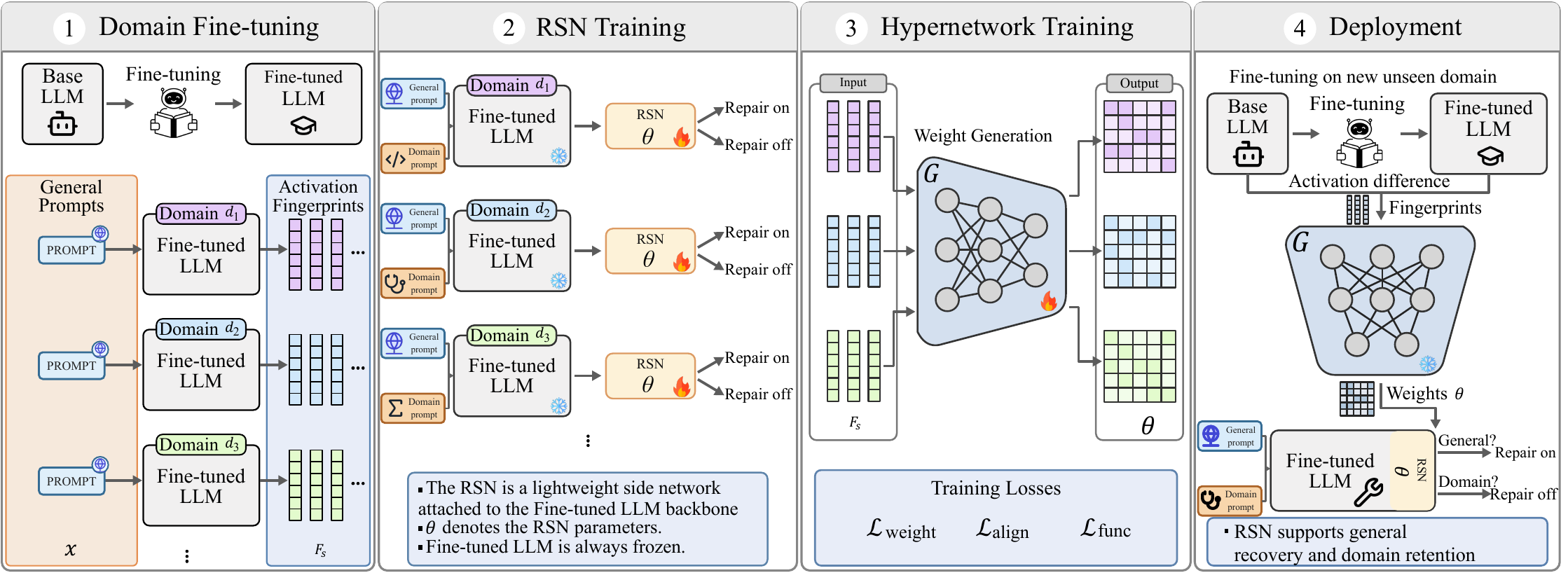}
\caption{Overview of \name{}. (1) Domain fine-tuning produces source models, whose activations on general prompts are used to construct fingerprints. (2) The gated \repairman{} is trained for each source model while keeping its backbone frozen. (3) A \hypernet{} learns to generate \repairman{} parameters from activation fingerprints using weight, representation-alignment, and functional supervision. (4) For a fine-tuned model from an unseen adaptation domain, the frozen \hypernet{} generates \repairman{} parameters without further optimization or target-domain data. The gate controls whether the repair is applied.}
\label{fig:pipeline}
\end{figure}

\name{} aims to recover general capabilities lost during domain fine-tuning while retaining domain performance. Figure~\ref{fig:pipeline} gives an overview of the framework. The framework combines an \repairman{}, a gated side network that corrects the frozen model's final hidden states, with \hypernet{}, which predicts the repair parameters from the model's activation fingerprint. We train \hypernet{} using fingerprints paired with \repairman{} parameters learned for source models. For a model from an unseen domain, the trained generator produces a repair without further optimization or examples from that domain. The following sections introduce the problem, the two networks, and the training and deployment procedure.

\subsection{Problem Formulation}
\label{sec:problem}

Let \(M_0\) denote a pretrained language model. Fine-tuning \(M_0\) on domain \(d\) produces the model \(M_d\). This adaptation can improve domain performance while reducing performance on general tasks. Our goal is to recover general capabilities and retain the domain capabilities acquired during fine-tuning. All parameters of \(M_d\) remain frozen during repair.

We seek a \hypernet{} \(G\) that generates an \repairman{} for a fine-tuned model from a potentially unseen domain. Given the model's activation fingerprint \(\mathbf{F}_d\), parameter generation and prompt gating are expressed as
\begin{equation}
\widehat{\theta}_d=G(\mathbf{F}_d),
\qquad
\lambda(x)=f_{\hat{\theta}_d}(x;M_d).
\label{eq:repair-generation}
\end{equation}
Here \(\mathbf{F}_d\) summarizes how \(M_d\)'s activations differ from those of \(M_0\) on fixed unlabeled general prompts. The generated parameters \(\widehat{\theta}_d\) specify the complete \repairman{}, including its correction network and prompt gate. The function \(f_{\hat{\theta}_d}\) denotes its gate computation using prompt \(x\) and the frozen model's activations. Its output \(\lambda(x)\in[0,1]\) controls the strength of the correction described in Section~\ref{sec:sidenet}. Each generator is trained on source models and applied to held-out models from domains excluded from that training. These models share the same pretrained backbone \(M_0\).

\subsection{Repairing Side Net}
\label{sec:sidenet}

\suppressfloats[t]
\begin{figure}[t]
\centering
\includegraphics[
  width=\linewidth
]{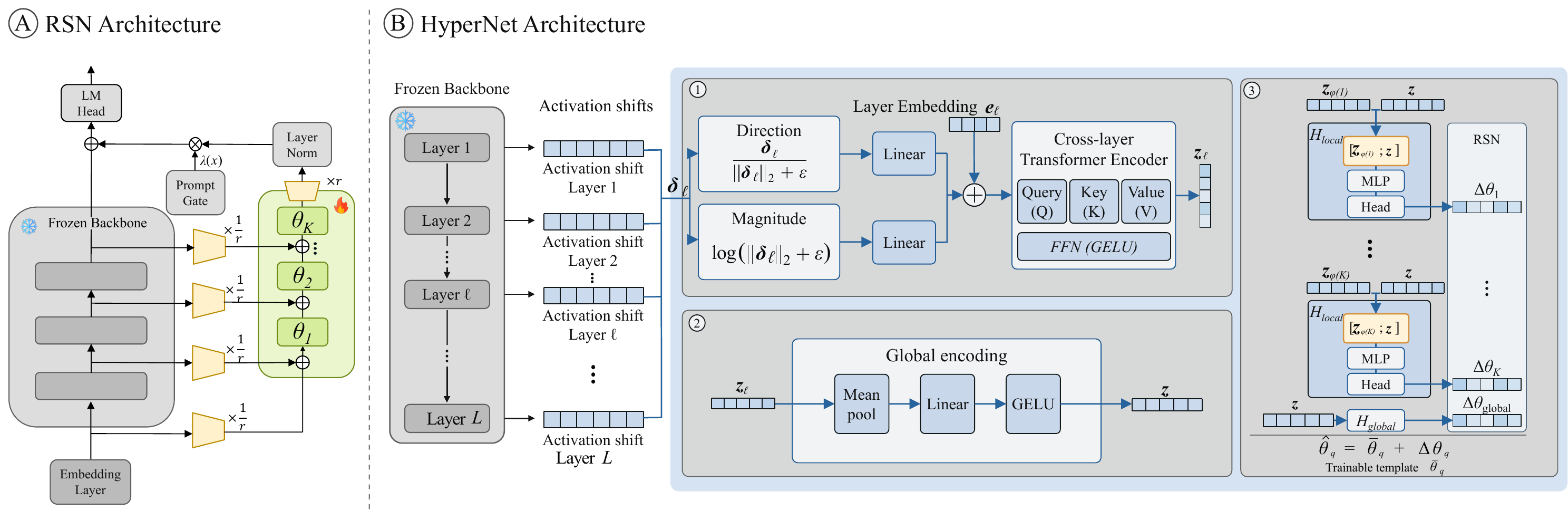}
\caption{Architectures of the \repairman{} and \hypernet{}. (A) The \repairman{} processes intermediate states from the frozen backbone. The prompt-conditioned gate \(\lambda(x)\) scales its final output before addition to the backbone state. The backbone and language-model head remain frozen throughout repair training. (B) \hypernet{} encodes layerwise activation shifts with a Transformer and pools the encodings into a global representation. It adds generated local and global updates to a trainable shared template.}
\label{fig:architecture}
\end{figure}

The \repairman{} attaches to the frozen backbone and produces a corrective residual together with a prompt gate. We adopt the side tuning technique to modify the backbone's final hidden states \citep{zhang2020sidetuning, sung2022lst}. Figure~\ref{fig:architecture}A shows how the correction enters the model's predictions.

\paragraph{Side Transformer.}
The \repairman{} uses \(K\) Transformer layers with hidden dimension \(h_s=h/r\), where \(h\) is the backbone hidden size and \(r\) is the reduction factor. We index the backbone layers by \(\ell=1,\ldots,L\) and the side layers by \(k=1,\ldots,K\). The initial side state \(\mathbf{h}^{\mathrm{side}}_0\) is obtained by projecting the backbone's input embeddings from \(h\) to \(h_s\) dimensions.

\paragraph{Ladder Connections.}
Each side layer combines a projected backbone hidden state with the output of the preceding side layer. Omitting projection biases, the input to side layer \(k\) is
\begin{equation}
\mathbf{h}^{\mathrm{in}}_k
=\mu_k\mathbf{W}^{\mathrm{down}}_k\mathbf{h}^{M_d}_{\phi(k)}
+(1-\mu_k)\mathbf{h}^{\mathrm{side}}_{k-1}.
\label{eq:sidenet-path}
\end{equation}
Here \(\phi:\{1,\ldots,K\}\to\{1,\ldots,L\}\) maps each side layer to its connected backbone layer, and \(\mathbf{W}^{\mathrm{down}}_k\in\mathbb{R}^{h_s\times h}\) projects that backbone layer's hidden state to the side-network width. The learned scalar \(\mu_k\in(0,1)\) controls the contribution of the backbone state. The side Transformer block maps \(\mathbf{h}^{\mathrm{in}}_k\) to its output \(\mathbf{h}^{\mathrm{side}}_k\). 

\paragraph{Selective Repair.}
After the final side layer, we project its output from \(h_s\) to \(h\) and normalize it to obtain the correction \(\mathbf{h}^{\mathrm{side}}_{\mathrm{final}}\). We average these states over prompt tokens and pass the result through a small multilayer perceptron (MLP) and sigmoid to produce \(\lambda(x)\). For each token, we add the gated correction to the backbone's final hidden state \(\mathbf{h}^{M_d}_{\mathrm{final}}\) before the frozen language-model head:
\begin{equation}
\mathbf{h}^{\mathrm{repaired}}_{\mathrm{final}}
=\mathbf{h}^{M_d}_{\mathrm{final}}+\lambda(x)\mathbf{h}^{\mathrm{side}}_{\mathrm{final}}.
\label{eq:gated-repair}
\end{equation}
The gate should be near one on general prompts to apply the correction. On domain prompts, it should be near zero to preserve the fine-tuned model's behavior. When the gate is zero, the correction is disabled and the model produces the original fine-tuned predictions. The gate depends only on the prompt and is shared when scoring answer candidates or generating a response. During training, we use the unthresholded sigmoid gate.

\subsection{Activation-Conditioned Hypernetwork}
\label{sec:fingerprint-generator}
\label{sec:parameter-generation}

\hypernet{} generates the parameters of an RSN from the activation fingerprint of a fine-tuned model (Figure~\ref{fig:architecture}B). We encode activation shifts by direction and magnitude and use low-rank factors to generate matrix updates.

\paragraph{Activation Fingerprint Extraction.}
To compare models under the same inputs, we use a fixed group of \(N\) unlabeled general prompts \(\{x_i\}_{i=1}^{N}\). Let \(\mathbf{h}_{\ell}^{M}(x)\) denote model \(M\)'s raw hidden state at decoder layer \(\ell\), evaluated at the final token of prompt \(x\). We measure the mean activation change relative to the pretrained model:
\begin{equation}
\boldsymbol{\delta}_\ell
=\frac{1}{N}\sum_{i=1}^{N}
\left[\mathbf{h}_{\ell}^{M_d}(x_i)-\mathbf{h}_{\ell}^{M_0}(x_i)\right].
\label{eq:fingerprint-shift}
\end{equation}
Using general prompts allows us to characterize a new model without data from its domain. 

\paragraph{Direction-Magnitude Decomposition.}
We represent each layer's activation change by its direction and log magnitude:
\begin{equation}
\mathbf{d}_\ell
=\frac{\boldsymbol{\delta}_\ell}
{\lVert\boldsymbol{\delta}_\ell\rVert_2+\epsilon},
\qquad
m_\ell
=\log\!\left(\lVert\boldsymbol{\delta}_\ell\rVert_2+\epsilon\right).
\label{eq:fingerprint}
\end{equation}
Here \(\epsilon\) is a small constant for numerical stability. The fingerprint \(\mathbf{F}_d\) collects \(\mathbf{d}_\ell\) and \(m_\ell\) across all backbone layers. A Transformer encodes these features together with learned layer embeddings, producing a local fingerprint encoding \(\mathbf{z}_\ell\) for each backbone layer. Pooling the local encodings gives the global fingerprint encoding \(\mathbf{z}\). 

\paragraph{Layerwise Factorized Generation.}
\hypernet{} generates \repairman{} parameters as a shared template plus an update conditioned on the fine-tuned model:
\begin{equation}
\widehat{\theta}_d=\bar{\theta}+\Delta\theta_d.
\label{eq:parameter-generation}
\end{equation}
The trainable template \(\bar{\theta}\) is shared across models. The update \(\Delta\theta_d\) depends on the fine-tuned model's fingerprint. For side layer \(k\), we generate its update from the corresponding local encoding and the global encoding:
\begin{equation}
\Delta\theta_k=H_{\mathrm{local}}\!\left(\mathbf{z}_{\phi(k)},\mathbf{z}\right).
\label{eq:local-parameter-generation}
\end{equation}
Here \(\mathbf{z}_{\phi(k)}\) is the local fingerprint encoding of the backbone layer connected to side layer \(k\). The local generator \(H_{\mathrm{local}}\) concatenates this local encoding with the global encoding \(\mathbf{z}\) and processes them with an MLP followed by parameter heads. The MLP and heads for each parameter role share weights across side layers. Matrix updates are products of low-rank factors predicted by the heads, reducing the number of generated values. Vector and scalar updates are predicted directly.

\paragraph{Global Parameter Generation.}
A separate generator \(H_{\mathrm{global}}\) takes \(\mathbf{z}\) and produces updates for the output projection, output normalization, and prompt-gate network. All updates are added to the corresponding shared template parameters. 

\subsection{Training and Deployment}
\label{sec:hypernet-training}

We first train an \repairman{} separately for each frozen source model. These trained networks then supply the parameter targets used to train \hypernet{}.

\paragraph{Source \repairman{} Training.}
For each source model \(M_d\), we train an \repairman{} and use its learned parameters \(\theta_d^\star\) as supervision targets. Labeled general examples teach it to correct predictions. General and unlabeled domain prompts provide gate supervision, with targets one and zero, respectively. Let \(\mathcal{L}_{\mathrm{general}}\) denote cross-entropy over candidate answers for discriminative tasks or mean response-token cross-entropy for generative tasks. The training objective is
\begin{equation}
\mathcal{L}_{\mathrm{func}}
=\mathcal{L}_{\mathrm{general}}
+\mathcal{L}_{\mathrm{gate}}.
\label{eq:functional-loss}
\end{equation}
Here \(\mathcal{L}_{\mathrm{gate}}\) is the binary cross-entropy loss that supervises the prompt gate on general and domain inputs. Domain answers are not required because domain prompts contribute only gate supervision. 

\paragraph{Hypernetwork Training.}
Each source model supplies a fingerprint \(\mathbf{F}_d\) and ground-truth RSN parameters \(\theta_d^\star\). We train \hypernet{} to reconstruct these parameters and optimize the predictions made with the generated \repairman{}. The combined objective is
\begin{equation}
\mathcal{L}_{\mathrm{hyper}}
=
\beta_{\mathrm{weight}}\mathcal{L}_{\mathrm{weight}}
+
\beta_{\mathrm{align}}\mathcal{L}_{\mathrm{align}}
+
\beta_{\mathrm{func}}\mathcal{L}_{\mathrm{func}},
\label{eq:hypernet-loss}
\end{equation}
where \(\mathcal{L}_{\mathrm{weight}}\) is the mean-squared error between the generated and ground-truth RSN weights ( \(\widehat{\theta}_d\) and \(\theta_d^\star\) ). \(\mathcal L_{\mathrm{align}}\) is the mean-squared error between the normalized gate-input features of the generated and ground-truth RSNs. 

The functional loss \(\mathcal{L}_{\mathrm{func}}\) applies Equation~\ref{eq:functional-loss} to the generated \repairman{} attached to its frozen source model. Because the prompt gate reads the correction states, errors in generated RSN parameters can affect both the correction and the gating decision. Representation alignment constrains the features used by the gate, complementing weight reconstruction and functional supervision. Training updates only \hypernet{}, including the shared template. The backbones and ground-truth \repairman{}s remain frozen. 

\paragraph{Deployment.}
For a held-out \(M_d\), we compute its fingerprint using the fixed general prompts, evaluate \hypernet{} once, and attach the generated \repairman{}. The repair modifies predictions through Equation~\ref{eq:gated-repair}, using the generated prompt gate to control its strength. Generating this repair requires no additional optimization or examples from the held-out model's domain.

\section{Experiments}
\label{sec:experiments}

\subsection{Experimental Setup}
\label{sec:experimental-setup}

\paragraph{Models and Domains.}
\label{sec:models-data}
We evaluate the base checkpoints of Qwen2.5-7B \citep{qwen2024qwen25} and Llama-3.1-8B \citep{grattafiori2024llama3} on 12 domains spanning classification, question answering, semantic parsing, and code summarization. We treat each dataset--task setting as an adaptation domain. For each backbone and domain, we fine-tune a LoRA adapter \citep{hu2022lora}, which remains active during repair.

\paragraph{General Benchmarks.}
We measure general capabilities on 15 general benchmarks: 12 discriminative tasks and three generative tasks, MATH, HumanEval, and MBPP. Domain-task scores measure retention of the capabilities acquired during fine-tuning. We report individual general-benchmark scores and their unweighted mean as the average general score. 

\paragraph{Evaluation Protocol.}
\label{sec:folds}
We use four splits, each holding out three of the 12 domains.
For each backbone and split, we train a separate \hypernet{} on the remaining nine domains and evaluate its generated \repairman{}s on the three held-out models. The main comparison uses the default split (Split D), which holds out LEDGAR-Top10, SNIPS-7, and DROP. 

\paragraph{Parameter and inference overhead.}
The \repairman{} adds 9.57M parameters to Qwen2.5-7B and 10.56M to Llama-3.1-8B, approximately \(0.13\%\) of each backbone. For each adapted model, repair generation requires fingerprint extraction on 150 unlabeled general prompts and one \hypernet{} forward pass. The generated parameters are reused across subsequent requests, while the \repairman{} runs at every decoding step.

\subsection{Main Results}
\label{sec:results}

\begin{table}[!t]
\centering
\caption{General benchmark test scores for Qwen2.5-7B and Llama-3.1-8B on Split D. Base is shared across held-out domains within each backbone. \name{} uses \repairman{}s generated by a \hypernet{} trained on the other nine domains.}
\label{tab:general-results}
\fontsize{7}{8.5}\selectfont
\setlength{\tabcolsep}{0.2pt}
\renewcommand{\arraystretch}{1.0}
\begin{tabular*}{\linewidth}{@{\extracolsep{\fill}}l*{15}{>{\centering\arraybackslash}p{17pt}}@{}}
\toprule
{\fontsize{8}{9.3}\selectfont\textbf{Method}} & \makebox[\linewidth][l]{\rotatebox{60}{\fontsize{7.5}{9}\selectfont\textbf{ARC-Challenge}}} & \makebox[\linewidth][l]{\rotatebox{60}{\fontsize{7.5}{9}\selectfont\textbf{HellaSwag}}} & \makebox[\linewidth][l]{\rotatebox{60}{\fontsize{7.5}{9}\selectfont\textbf{MMLU}}} & \makebox[\linewidth][l]{\rotatebox{60}{\fontsize{7.5}{9}\selectfont\textbf{OpenBookQA}}} & \makebox[\linewidth][l]{\rotatebox{60}{\fontsize{7.5}{9}\selectfont\textbf{PIQA}}} & \makebox[\linewidth][l]{\rotatebox{60}{\fontsize{7.5}{9}\selectfont\textbf{SciQ}}} & \makebox[\linewidth][l]{\rotatebox{60}{\fontsize{7.5}{9}\selectfont\textbf{WinoGrande}}} & \makebox[\linewidth][l]{\rotatebox{60}{\fontsize{7.5}{9}\selectfont\textbf{ANLI-R1}}} & \makebox[\linewidth][l]{\rotatebox{60}{\fontsize{7.5}{9}\selectfont\textbf{SocialIQA}}} & \makebox[\linewidth][l]{\rotatebox{60}{\fontsize{7.5}{9}\selectfont\textbf{LogiQA}}} & \makebox[\linewidth][l]{\rotatebox{60}{\fontsize{7.5}{9}\selectfont\textbf{WiC}}} & \makebox[\linewidth][l]{\rotatebox{60}{\fontsize{7.5}{9}\selectfont\textbf{COPA}}} & \makebox[\linewidth][l]{\rotatebox{60}{\fontsize{7.5}{9}\selectfont\textbf{MATH}}} & \makebox[\linewidth][l]{\rotatebox{60}{\fontsize{7.5}{9}\selectfont\textbf{HumanEval}}} & \makebox[\linewidth][l]{\rotatebox{60}{\fontsize{7.5}{9}\selectfont\textbf{MBPP}}} \\
\midrule
\multicolumn{16}{@{}l}{{\fontsize{9}{10.5}\selectfont\textbf{\textit{Qwen2.5-7B}}}} \\
\midrule
{\fontsize{8}{9.3}\selectfont Base} & 51.4 & 78.8 & 71.8 & 47.4 & 79.8 & 95.2 & 73.1 & 53.0 & 54.7 & 35.8 & 57.8 & 91 & 39.3 & 56.1 & 63.8 \\
\midrule
\multicolumn{16}{l}{{\fontsize{8}{9.3}\selectfont \textbf{LEDGAR-Top10}}} \\
\addlinespace[1pt]
{\fontsize{8}{9.3}\selectfont Fine-tuned} & 53.8 & 73.1 & 55.0 & 46.0 & 79.4 & 88.1 & 70.8 & 48.5 & 49.0 & 33.6 & 50.0 & 88 & 3.6 & 22.0 & 28.8 \\
{\fontsize{8}{9.3}\selectfont CorDA-KPM} & 45.7 & 65.7 & 24.0 & 44.2 & 75.7 & 78.1 & 69.6 & 33.0 & 46.8 & 27.0 & 50.0 & 90 & 0.0 & 4.3 & 0.0 \\
{\fontsize{8}{9.3}\selectfont FAPM} & 55.4 & 75.0 & \textbf{63.1} & 48.0 & 80.5 & 91.3 & 71.9 & 49.0 & \textbf{51.1} & 33.3 & 50.0 & 89 & 28.5 & \textbf{34.8} & \textbf{44.0} \\
{\fontsize{8}{9.3}\selectfont FLOW} & 27.6 & 41.5 & 28.9 & 31.8 & 60.4 & 50.5 & 50.7 & 31.4 & 37.9 & 29.3 & \textbf{50.2} & 57 & 0.3 & 9.8 & 25.0 \\
{\fontsize{8}{9.3}\selectfont \shortstack[l]{\textbf{\name{}}}} & \textbf{65.4} & \textbf{84.3} & 62.9 & \textbf{51.2} & \textbf{84.8} & \textbf{93.3} & \textbf{75.4} & \textbf{54.7} & \textbf{51.1} & \textbf{34.1} & 50.0 & \textbf{95} & \textbf{39.4} & \textbf{34.8} & 41.8 \\
\midrule
\multicolumn{16}{l}{{\fontsize{8}{9.3}\selectfont \textbf{SNIPS-7}}} \\
\addlinespace[1pt]
{\fontsize{8}{9.3}\selectfont Fine-tuned} & 44.8 & 70.3 & 61.8 & 41.8 & 75.1 & 87.3 & 68.7 & 37.1 & 38.0 & 31.0 & 50.3 & 85 & 0.0 & 1.8 & 0.0 \\
{\fontsize{8}{9.3}\selectfont CorDA-KPM} & 32.6 & 53.2 & 24.0 & 38.2 & 67.8 & 71.5 & 62.4 & 32.2 & 37.6 & 27.5 & 50.0 & 85 & 0.0 & 0.0 & 0.0 \\
{\fontsize{8}{9.3}\selectfont FAPM} & 46.7 & 73.1 & 68.2 & 42.4 & 76.6 & 92.3 & 69.8 & \textbf{42.1} & 39.3 & 31.6 & 52.4 & 82 & 0.0 & 14.0 & 12.4 \\
{\fontsize{8}{9.3}\selectfont FLOW} & 53.0 & 74.6 & \textbf{69.6} & 43.6 & 77.7 & \textbf{95.6} & 70.3 & 41.7 & \textbf{46.5} & \textbf{33.8} & \textbf{61.8} & 88 & 9.8 & 32.3 & \textbf{27.6} \\
{\fontsize{8}{9.3}\selectfont \shortstack[l]{\textbf{\name{}}}} & \textbf{55.7} & \textbf{80.9} & 64.8 & \textbf{46.6} & \textbf{78.1} & 89.5 & \textbf{71.3} & \textbf{42.1} & 38.0 & 30.9 & 50.3 & \textbf{90} & \textbf{35.8} & \textbf{36.0} & 23.2 \\
\midrule
\multicolumn{16}{l}{{\fontsize{8}{9.3}\selectfont \textbf{DROP}}} \\
\addlinespace[1pt]
{\fontsize{8}{9.3}\selectfont Fine-tuned} & 59.9 & 76.1 & 64.8 & 46.8 & 79.8 & \textbf{97.3} & 75.1 & 47.0 & 54.9 & 44.9 & 65.8 & 90 & 33.4 & 51.8 & 58.8 \\
{\fontsize{8}{9.3}\selectfont CorDA-KPM} & 50.8 & 66.1 & 43.6 & 44.2 & 76.6 & 96.5 & 67.8 & 36.1 & 51.7 & 27.5 & 52.7 & 90 & 0.0 & 8.5 & 10.6 \\
{\fontsize{8}{9.3}\selectfont FAPM} & 61.2 & 76.6 & \textbf{66.4} & 45.2 & 80.5 & \textbf{97.3} & 74.4 & 47.6 & 54.9 & \textbf{46.1} & \textbf{66.3} & 91 & 39.7 & 50.6 & \textbf{60.6} \\
{\fontsize{8}{9.3}\selectfont FLOW} & 61.3 & 76.5 & 66.2 & 47.8 & 79.9 & 97.0 & 74.0 & 44.9 & 55.6 & 43.2 & 59.4 & 91 & 25.7 & \textbf{53.0} & \textbf{60.6} \\
{\fontsize{8}{9.3}\selectfont \shortstack[l]{\textbf{\name{}}}} & \textbf{70.4} & \textbf{87.6} & \textbf{66.4} & \textbf{52.8} & \textbf{84.2} & \textbf{97.3} & \textbf{80.7} & \textbf{52.3} & \textbf{59.3} & 42.7 & 62.7 & \textbf{95} & \textbf{40.1} & 52.4 & 56.8 \\
\midrule[0.8pt]
\multicolumn{16}{@{}l}{{\fontsize{9}{10.5}\selectfont\textbf{\textit{Llama-3.1-8B}}}} \\
\midrule
{\fontsize{8}{9.3}\selectfont Base} & 54.7 & 79.3 & 64.0 & 45.4 & 81.0 & 95.6 & 74.4 & 35.1 & 48.4 & 31.0 & 51.1 & 92 & 19.9 & 36.6 & 47.6 \\
\midrule
\multicolumn{16}{l}{{\fontsize{8}{9.3}\selectfont \textbf{LEDGAR-Top10}}} \\
\addlinespace[1pt]
{\fontsize{8}{9.3}\selectfont Fine-tuned} & 40.1 & 62.6 & 28.1 & 42.8 & 74.0 & 69.0 & 71.0 & 36.2 & 40.7 & \textbf{31.2} & \textbf{50.0} & 86 & 0.0 & 0.0 & 0.0 \\
{\fontsize{8}{9.3}\selectfont CorDA-KPM} & 48.8 & 73.6 & 25.5 & 43.6 & 79.0 & \textbf{88.7} & 73.6 & 33.3 & \textbf{48.0} & 29.5 & \textbf{50.0} & 90 & 0.0 & \textbf{22.6} & \textbf{20.2} \\
{\fontsize{8}{9.3}\selectfont FAPM} & 41.5 & 67.1 & \textbf{31.3} & 42.6 & 76.2 & 75.6 & 71.7 & 36.3 & 41.8 & 30.7 & \textbf{50.0} & 88 & 0.0 & 0.0 & 0.0 \\
{\fontsize{8}{9.3}\selectfont FLOW} & 39.8 & 69.2 & 24.2 & 43.8 & 76.5 & 81.3 & 73.4 & 32.5 & 42.8 & \textbf{31.2} & \textbf{50.0} & \textbf{91} & 0.0 & 1.8 & 0.0 \\
{\fontsize{8}{9.3}\selectfont \shortstack[l]{\textbf{\name{}}}} & \textbf{49.5} & \textbf{75.9} & 28.4 & \textbf{44.6} & \textbf{79.2} & 81.3 & \textbf{76.8} & \textbf{36.9} & 43.7 & 29.6 & \textbf{50.0} & \textbf{91} & \textbf{10.8} & 14.6 & 17.6\\
\midrule
\multicolumn{16}{l}{{\fontsize{8}{9.3}\selectfont \textbf{SNIPS-7}}} \\
\addlinespace[1pt]
{\fontsize{8}{9.3}\selectfont Fine-tuned} & 38.1 & 68.3 & 23.5 & 41.0 & 75.9 & 76.7 & 70.7 & 33.2 & 40.3 & 29.3 & 49.8 & 88 & 0.0 & 12.2 & 1.2 \\
{\fontsize{8}{9.3}\selectfont CorDA-KPM} & \textbf{47.7} & 76.4 & 34.0 & 43.4 & 78.6 & \textbf{93.4} & 70.9 & 31.2 & \textbf{43.7} & 29.6 & 50.0 & 89 & 0.1 & 19.5 & 16.4 \\
{\fontsize{8}{9.3}\selectfont FAPM} & 39.4 & 70.5 & 23.7 & 41.0 & 77.1 & 80.2 & 71.1 & 33.2 & 41.7 & 30.1 & 49.5 & 89 & 0.0 & 16.5 & 6.2 \\
{\fontsize{8}{9.3}\selectfont FLOW} & 41.2 & 68.2 & \textbf{34.8} & 40.4 & 76.6 & 84.2 & 70.0 & 32.6 & 39.6 & 27.6 & 50.8 & 86 & 0.0 & 3.7 & 0.0 \\
{\fontsize{8}{9.3}\selectfont \shortstack[l]{\textbf{\name{}}}} & \textbf{47.7} & \textbf{76.7} & 28.1 & \textbf{47.2} & \textbf{79.0} & 81.5 & \textbf{75.4} & \textbf{41.4} & 42.4 & \textbf{30.6} & \textbf{51.1} & \textbf{90} & \textbf{9.4} & \textbf{20.7} & \textbf{16.8} \\
\midrule
\multicolumn{16}{l}{{\fontsize{8}{9.3}\selectfont \textbf{DROP}}} \\
\addlinespace[1pt]
{\fontsize{8}{9.3}\selectfont Fine-tuned} & 55.2 & 76.2 & 33.2 & 47.2 & 79.8 & 96.3 & 74.5 & 32.1 & 51.5 & 27.5 & 53.1 & 92 & 0.0 & 13.4 & 25.4 \\
{\fontsize{8}{9.3}\selectfont CorDA-KPM} & 54.9 & 74.7 & 35.0 & 42.8 & 79.2 & 96.9 & 72.1 & 31.6 & 51.0 & 27.8 & 50.3 & 88 & 0.0 & 20.1 & 18.6 \\
{\fontsize{8}{9.3}\selectfont FAPM} & 56.1 & 76.9 & 35.8 & 47.0 & 80.3 & 96.4 & 74.7 & 32.4 & 51.9 & 28.3 & 54.7 & 92 & 0.0 & 17.1 & 30.2 \\
{\fontsize{8}{9.3}\selectfont FLOW} & 55.2 & 76.1 & 29.6 & 46.0 & 79.9 & 96.9 & 75.0 & 33.0 & 50.9 & \textbf{28.6} & 50.0 & 92 & 0.0 & 15.9 & 33.4 \\
{\fontsize{8}{9.3}\selectfont \shortstack[l]{\textbf{\name{}}}} & \textbf{60.6} & \textbf{83.2} & \textbf{38.5} & \textbf{53.4} & \textbf{83.1} & \textbf{97.2} & \textbf{79.8} & \textbf{41.8} & \textbf{53.0} & 28.1 & \textbf{61.0} & \textbf{93} & \textbf{8.2} & \textbf{22.6} & \textbf{34.4} \\
\bottomrule
\end{tabular*}
\end{table}

\begin{figure}[!t]
\centering
\includegraphics[width=\linewidth]{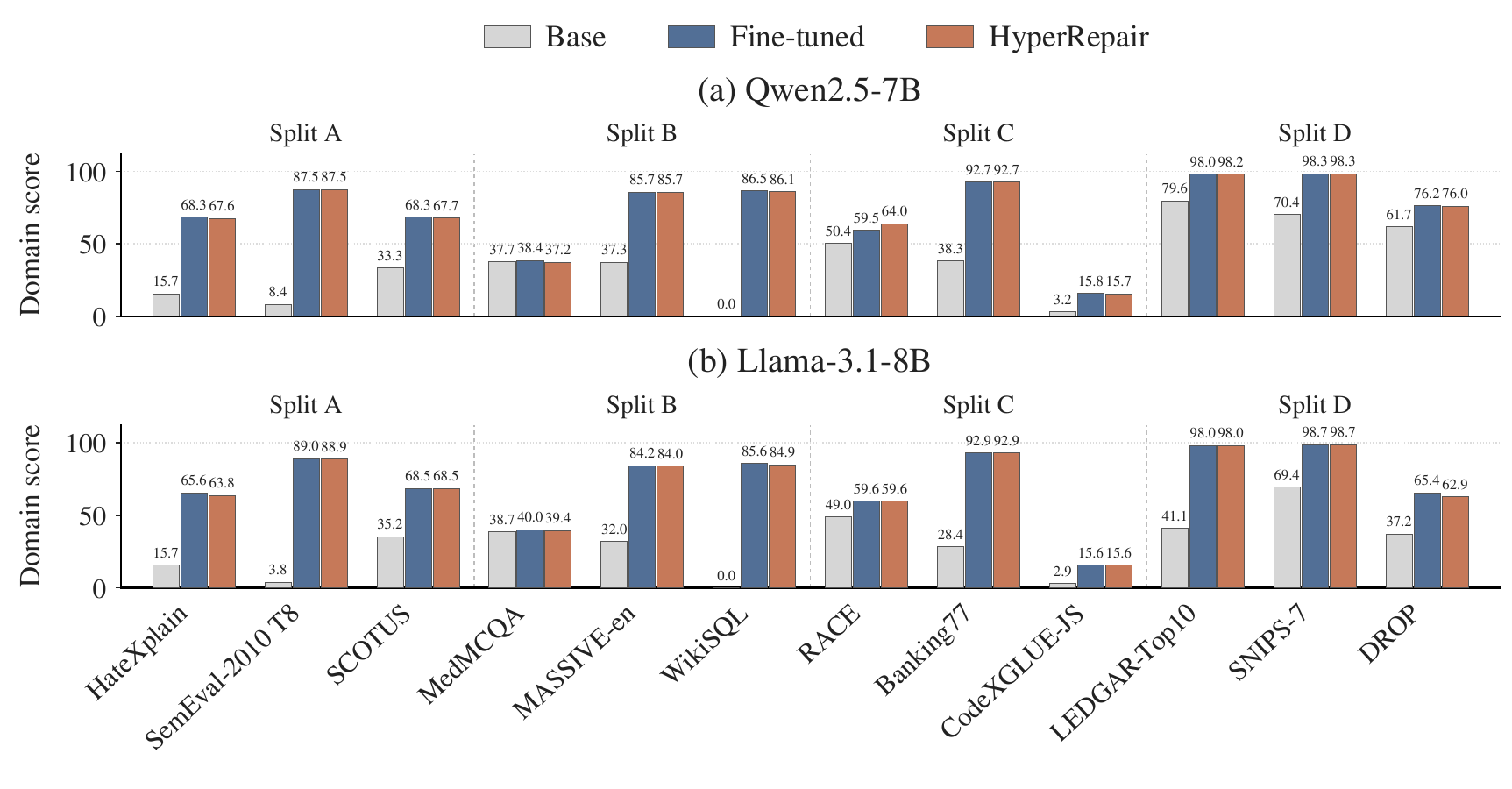}
\caption{Domain test performance across the 12 held-out adaptation domains. HyperRepair generally retains the scores of the fine-tuned models on both backbones.}
\label{fig:domain-results}
\end{figure}

\label{sec:results-lora-premise}
\label{sec:results-supervised}
\label{sec:results-held-out}
\label{sec:results-all-domains}

\paragraph{General recovery and domain retention.}
HyperRepair improves the average general score of all six held-out models in Table~\ref{tab:general-results}, reaching 93.4\% of the base models' average general score. Across the 12 held-out domains, domain performance remains close to that of the fine-tuned models (Figure~\ref{fig:domain-results}); on Split D, HyperRepair retains 99.5\% of their average domain score. The gating ablation on Qwen2.5-7B supports the role of selective repair: keeping the repair always active raises the average general score but reduces the average domain score (Section~\ref{sec:ablations}).

\paragraph{Comparison with baselines.}
HyperRepair achieves the highest average general score on both backbones among the compared methods (Table~\ref{tab:general-results}). Averaged over 15 benchmarks and three held-out domains, it reaches 61.2 on Qwen2.5-7B and 51.2 on Llama-3.1-8B, compared with 57.0 and 48.7 for the strongest respective baselines, FAPM and CorDA-KPM. Because these models come from adaptation domains excluded from HyperNet training, the gains demonstrate that the learned repair mechanism generalizes beyond its source models. Activation fingerprints let HyperNet tailor this mechanism to the changes induced by fine-tuning in each target model. On Qwen2.5-7B, a shared RSN without fingerprint conditioning reduces the average general score (Section~\ref{sec:ablations}), supporting the value of model-specific repairs.

\subsection{Ablation Studies}
\label{sec:ablations}
We conduct ablation studies on Qwen2.5-7B, reporting the changes in average general score and domain score, averaged across the three held-out domains in Split D. The full system achieves the best overall performance. Removing any component degrades at least one metric.
\begin{figure}[!ht]
\centering
\includegraphics[width=\linewidth]{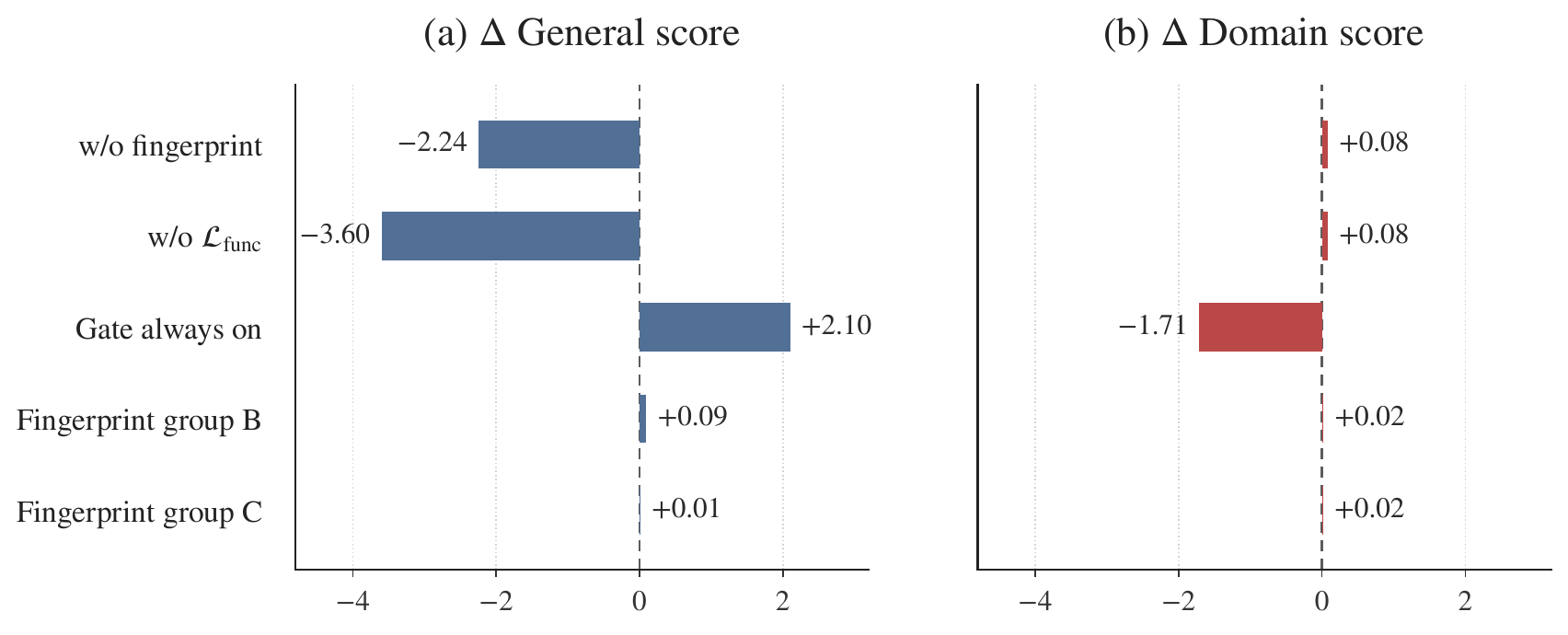}
\caption{Ablation study on Qwen2.5-7B across the three held-out domains in Split D. Values are changes in general and domain scores relative to the full model, measured in score points.}
\label{fig:ablations}
\end{figure}

\paragraph{Loss components.}
\label{sec:ablation-loss}
Removing $\mathcal{L}_{\mathrm{func}}$ while retaining $\mathcal{L}_{\mathrm{weight}}$ and $\mathcal{L}_{\mathrm{align}}$ reduces general recovery. Matching source RSN weights and features alone is less effective than also supervising the predictions and gate of the generated repair.

\paragraph{Activation fingerprints.}
\label{sec:ablation-fingerprint}
We remove the fingerprint-conditioned updates and train only the shared RSN template, keeping the three losses and training budget unchanged. All held-out models use the same repair parameters, while prompt-conditioned gating is retained.
Using a shared RSN reduces the average general score by 2.24 points, with an increase of 0.08 points in domain score. This supports the value of tailoring repair parameters to individual fine-tuned models.

\paragraph{Fingerprint sampling sensitivity.}
\label{sec:ablation-fingerprint-sampling}
Aggregate scores change little when the fingerprint is recomputed from different prompts. We compare three disjoint groups of 150 general prompts with matched task composition, denoted A, B, and C. Group A is used during \hypernet{} training; evaluation substitutes B or C while holding the fine-tuned model and \hypernet{} fixed. These results support low sensitivity to the tested prompt substitutions. 

\paragraph{Prompt-conditioned gating.}
\label{sec:ablation-gate}
Keeping the repair always active ($\lambda(x)=1$) reduces domain performance, showing that applying corrections to every prompt can interfere with domain capabilities. These results highlight the importance of controlling when and how strongly the repair is applied.

\subsection{Qualitative Results}
\label{sec:qualitative-results}

Table~\ref{tab:main-recovery-cases} shows two representative recovery examples on Qwen2.5-7B. On ARC-Challenge, fine-tuning on HateXplain changes the correct prediction, while HyperRepair restores it. On MATH, fine-tuning on MASSIVE-en leads to an error in handling repeated letters, and HyperRepair recovers the correct answer of 1260. More examples are provided in \cref{app:qualitative-recovery}.

\begin{table}[H]
\centering
\caption{Recovery examples on Qwen2.5-7B for ARC-Challenge and MATH. HyperRepair restores the correct outputs after fine-tuning errors.}
\label{tab:main-recovery-cases}
\small
\setlength{\tabcolsep}{5pt}
\renewcommand{\arraystretch}{1.08}
\begin{tabularx}{\linewidth}{@{}*{3}{>{\raggedright\arraybackslash}X}@{}}
\toprule
\textbf{Base (correct)} & \textbf{Fine-tuned (incorrect)} & \textbf{\name{} (correct)} \\
\midrule
\multicolumn{3}{@{}p{\linewidth}@{}}{\textbf{ARC-Challenge (discriminative)}\hfill\textit{Adaptation: HateXplain (Split A)}}\\[3pt]
\multicolumn{3}{@{}p{\linewidth}@{}}{\textbf{Prompt.} An astronomer observes that a planet rotates faster after a meteorite impact. Which is the most likely effect of this increase in rotation?}\\[5pt]
Planetary days will become shorter. & Planetary gravity will become stronger. & Planetary days will become shorter. \\
\addlinespace[6pt]\midrule
\multicolumn{3}{@{}p{\linewidth}@{}}{\textbf{MATH (generative)}\hfill\textit{Adaptation: MASSIVE-en (Split B)}}\\[3pt]
\multicolumn{3}{@{}p{\linewidth}@{}}{\textbf{Prompt.} Determine the number of ways to arrange the letters of the word ELLIPSE.}\\[5pt]
The word ELLIPSE has 7 letters, with 2 L's and 2 E's. The number of ways to arrange the letters is $\frac{7!}{2!2!}=\boxed{1260}.$ & The word ELLIPSE has 7 letters, but the letter E appears twice. Therefore, the number of ways to arrange the letters is $\frac{7!}{2!}=\boxed{2520}.$ & The word ELLIPSE has 7 letters, with 2 L's and 2 E's.  The number of ways to arrange the letters is $\frac{7!}{2!2!}=\boxed{1260}$. \\
\bottomrule
\end{tabularx}
\end{table}

\section{Conclusion}
\label{sec:conclusion}
HyperRepair generates model-specific RSNs that recover general capabilities in frozen fine-tuned language models across diverse adaptation domains and two model families while maintaining strong domain performance. Two findings highlight its practical value: a hypernetwork trained on source models generates effective repairs for models from unseen adaptation domains without target-domain examples or additional optimization, and prompt-conditioned gating preserves domain capabilities that deteriorate when the repair is always active.


\bibliography{iclr2027_conference}

@string {S = "Science"}

@string {N = "Nature"}

@string {C = "Cell"}

@string {CVPR = "Conference on Computer Vision and Pattern Recognition (CVPR)"}

@string {ECCV = "European Conference on Computer Vision (ECCV)"}

@string {ICML = "International Conference on Machine Learning (ICML)"}

@string {ICLR = "International Conference on Learning Representations (ICLR)"}

@string {AAAI = "AAAI Conference on Artificial Intelligence (AAAI)"}

@string {IJCAI = "International Joint Conference on Artificial Intelligence (IJCAI)"}

@string {ACL = "Annual Meeting of the Association for Computational Linguistics (ACL)"}

@string {EMNLP = "Annual Conference on Empirical Methods in Natural Language Processing (EMNLP)"}

@string {EMNLP-IJCNLP = "Proceedings of the 2019 Conference on Empirical Methods in Natural Language Processing and the 9th International Joint Conference on Natural Language Processing (EMNLP-IJCNLP)"}

@inproceedings{springer2025overtrained,
  title = {Overtrained Language Models Are Harder to Fine-Tune},
  author = {Springer, Jacob Mitchell and Goyal, Sachin and Wen, Kaiyue and Kumar, Tanishq and Yue, Xiang and Malladi, Sadhika and Neubig, Graham and Raghunathan, Aditi},
  booktitle = {International Conference on Machine Learning (ICML)},
  year = {2025}
}

@inproceedings{wortsman2022wiseft,
  title = {Robust Fine-Tuning of Zero-Shot Models},
  author = {Wortsman, Mitchell and Ilharco, Gabriel and Kim, Jong Wook and Li, Mike and Kornblith, Simon and Roelofs, Rebecca and Lopes, Raphael Gontijo and Hajishirzi, Hannaneh and Farhadi, Ali and Namkoong, Hongseok and Schmidt, Ludwig},
  booktitle = {IEEE/CVF Conference on Computer Vision and Pattern Recognition (CVPR)},
  year = {2022}
}

@inproceedings{wortsman2022soups,
  title = {{Model soups}: averaging weights of multiple fine-tuned models improves accuracy without increasing inference time},
  author = {Wortsman, Mitchell and Ilharco, Gabriel and Gadre, Samir Yitzhak and Roelofs, Rebecca and Gontijo-Lopes, Raphael and Morcos, Ari S and Namkoong, Hongseok and Farhadi, Ali and Carmon, Yair and Kornblith, Simon and Schmidt, Ludwig},
  booktitle = {International Conference on Machine Learning (ICML)},
  year = {2022}
}

@inproceedings{hu2022lora,
  title = {{LoRA}: Low-Rank Adaptation of Large Language Models},
  author = {Hu, Edward J. and Shen, Yelong and Wallis, Phillip and Allen-Zhu, Zeyuan and Li, Yuanzhi and Wang, Shean and Wang, Lu and Chen, Weizhu},
  booktitle = {International Conference on Learning Representations (ICLR)},
  year = {2022}
}

@inproceedings{yang2024corda,
  title = {{CorDA}: Context-Oriented Decomposition Adaptation of Large Language Models for Task-Aware Parameter-Efficient Fine-tuning},
  author = {Yang, Yibo and Li, Xiaojie and Zhou, Zhongzhu and Song, Shuaiwen Leon and Wu, Jianlong and Nie, Liqiang and Ghanem, Bernard},
  booktitle = {Advances in Neural Information Processing Systems (NeurIPS)},
  year = {2024}
}

@inproceedings{zhang2020sidetuning,
  title = {{Side-Tuning}: A Baseline for Network Adaptation via Additive Side Networks},
  author = {Zhang, Jeffrey O. and Sax, Alexander and Zamir, Amir and Guibas, Leonidas and Malik, Jitendra},
  booktitle = {European Conference on Computer Vision (ECCV)},
  year = {2020}
}

@inproceedings{sung2022lst,
  title = {{LST}: {Ladder Side-Tuning} for Parameter and Memory Efficient Transfer Learning},
  author = {Sung, Yi-Lin and Cho, Jaemin and Bansal, Mohit},
  booktitle = {Advances in Neural Information Processing Systems (NeurIPS)},
  year = {2022}
}

@inproceedings{ha2017hypernetworks,
  title = {{HyperNetworks}},
  author = {David Ha and Andrew Dai and Quoc V. Le},
  booktitle = {International Conference on Learning Representations (ICLR)},
  year = {2017}
}

@inproceedings{zheng2025spurious,
  title = {Spurious Forgetting in Continual Learning of Language Models},
  author = {Zheng, Junhao and Cai, Xidi and Qiu, Shengjie and Ma, Qianli},
  booktitle = {International Conference on Learning Representations (ICLR)},
  year = {2025}
}

@inproceedings{jiang2025functionvectors,
  title = {Unlocking the Power of Function Vectors for Characterizing and Mitigating Catastrophic Forgetting in Continual Instruction Tuning},
  author = {Jiang, Gangwei and Jiang, Caigao and Li, Zhaoyi and Xue, Siqiao and Zhou, Jun and Song, Linqi and Lian, Defu and Wei, Ying},
  booktitle = {International Conference on Learning Representations (ICLR)},
  year = {2025}
}

@inproceedings{sanyal2025flow,
  title = {Upweighting Easy Samples in Fine-Tuning Mitigates Forgetting},
  author = {Sanyal, Sunny and Prairie, Hayden and Das, Rudrajit and Kavis, Ali and Sanghavi, Sujay},
  booktitle = {International Conference on Machine Learning (ICML)},
  year = {2025}
}

@inproceedings{huang2025fapm,
  title = {Mitigating Catastrophic Forgetting in Large Language Models with Forgetting-aware Pruning},
  author = {Huang, Wei and Cheng, Anda and Wang, Yinggui},
  booktitle = {Conference on Empirical Methods in Natural Language Processing (EMNLP)},
  year = {2025}
}

@inproceedings{lin2026talr,
  title = {{SFT} Doesn't Always Hurt General Capabilities: Revisiting Domain-Specific Fine-Tuning in {LLMs}},
  author = {Lin, Jiacheng and Wang, Zhongruo and Qian, Kun and Wang, Tian and Srinivasan, Arvind and Zeng, Hansi and Jiao, Ruochen and Zhou, Xie and Gesi, Jiri and Wang, Dakuo and Guo, Yufan and Zhong, Kai and Zhang, Weiqi and Sanghavi, Sujay and Chen, Changyou and Yun, Hyokun and Li, Lihong},
  booktitle = {International Conference on Learning Representations (ICLR)},
  year = {2026}
}

@article{aliev2026hypersafe,
  title = {{HyperSafe}: Inference-Time Safety Recovery for Fine-Tuned Language Models},
  author = {Aliev, Aznaur and Hinojosa, Carlos and Eldesokey, Abdelrahman and An, Bang and Ghanem, Bernard and Yang, Yibo},
  journal = {arXiv preprint arXiv:2607.11475},
  year = {2026}
}

@inproceedings{kotha2024implicit,
  title = {Understanding Catastrophic Forgetting in Language Models via Implicit Inference},
  author = {Kotha, Suhas and Springer, Jacob Mitchell and Raghunathan, Aditi},
  booktitle = {International Conference on Learning Representations (ICLR)},
  year = {2024}
}

@article{kirkpatrick2017ewc,
  title = {Overcoming catastrophic forgetting in neural networks},
  author = {Kirkpatrick, James and Pascanu, Razvan and Rabinowitz, Neil and Veness, Joel and Desjardins, Guillaume and Rusu, Andrei A. and Milan, Kieran and Quan, John and Ramalho, Tiago and Grabska-Barwinska, Agnieszka and Hassabis, Demis and Clopath, Claudia and Kumaran, Dharshan and Hadsell, Raia},
  journal = {Proceedings of the National Academy of Sciences},
  volume = {114},
  number = {13},
  pages = {3521--3526},
  year = {2017},
  doi = {10.1073/pnas.1611835114}
}

@inproceedings{li2016lwf,
  title = {Learning Without Forgetting},
  author = {Li, Zhizhong and Hoiem, Derek},
  booktitle = {European Conference on Computer Vision (ECCV)},
  year = {2016}
}

@inproceedings{zhang2024dissecting,
  title = {Dissecting learning and forgetting in language model finetuning},
  author = {Zhang, Xiao and Wu, Ji},
  booktitle = {International Conference on Learning Representations (ICLR)},
  year = {2024}
}

@inproceedings{sun2020lamol,
  title = {{LAMOL}: {LAnguage MOdeling} for Lifelong Language Learning},
  author = {Sun, Fan-Keng and Ho, Cheng-Hao and Lee, Hung-Yi},
  booktitle = {International Conference on Learning Representations (ICLR)},
  year = {2020}
}

@inproceedings{huang2024ssr,
  title = {Mitigating Catastrophic Forgetting in Large Language Models with Self-Synthesized Rehearsal},
  author = {Huang, Jianheng and Cui, Leyang and Wang, Ante and Yang, Chengyi and Liao, Xinting and Song, Linfeng and Yao, Junfeng and Su, Jinsong},
  booktitle = {Annual Meeting of the Association for Computational Linguistics (ACL)},
  year = {2024}
}

@inproceedings{harmon2026mapping,
  title = {Mapping Post-Training Forgetting in Language Models at Scale},
  author = {Harmon, Jackson and Hochlehnert, Andreas and Bethge, Matthias and Prabhu, Ameya},
  booktitle = {International Conference on Learning Representations (ICLR)},
  year = {2026}
}

@inproceedings{wang2025lines,
  title = {{LiNeS}: Post-training Layer Scaling Prevents Forgetting and Enhances Model Merging},
  author = {Wang, Ke and Dimitriadis, Nikolaos and Favero, Alessandro and Ortiz-Jimenez, Guillermo and Fleuret, Fran\c{c}ois and Frossard, Pascal},
  booktitle = {International Conference on Learning Representations (ICLR)},
  year = {2025}
}

@inproceedings{ilharco2023taskarithmetic,
  title = {Editing Models with Task Arithmetic},
  author = {Ilharco, Gabriel and Ribeiro, Marco Tulio and Wortsman, Mitchell and Gururangan, Suchin and Schmidt, Ludwig and Hajishirzi, Hannaneh and Farhadi, Ali},
  booktitle = {International Conference on Learning Representations (ICLR)},
  year = {2023}
}

@inproceedings{yadav2023ties,
  title = {{TIES}-Merging: Resolving Interference When Merging Models},
  author = {Yadav, Prateek and Tam, Derek and Choshen, Leshem and Raffel, Colin and Bansal, Mohit},
  booktitle = {Advances in Neural Information Processing Systems (NeurIPS)},
  year = {2023}
}

@inproceedings{huang2024emr,
  title = {{EMR-Merging}: Tuning-Free High-Performance Model Merging},
  author = {Huang, Chenyu and Ye, Peng and Chen, Tao and He, Tong and Yue, Xiangyu and Ouyang, Wanli},
  booktitle = {Advances in Neural Information Processing Systems (NeurIPS)},
  year = {2024}
}

@inproceedings{yu2024dare,
  title = {Language Models are {Super Mario}: Absorbing Abilities from Homologous Models as a Free Lunch},
  author = {Yu, Le and Yu, Bowen and Yu, Haiyang and Huang, Fei and Li, Yongbin},
  booktitle = {International Conference on Machine Learning (ICML)},
  year = {2024}
}

@misc{qwen2024qwen25,
  title = {{Qwen2.5}: A Party of Foundation Models},
  author = {{Qwen Team}},
  year = {2024},
  url = {https://qwenlm.github.io/blog/qwen2.5/}
}

@article{grattafiori2024llama3,
  title = {The {Llama 3} Herd of Models},
  author = {Grattafiori, Aaron and Dubey, Abhimanyu and Jauhri, Abhinav and others},
  journal = {arXiv preprint arXiv:2407.21783},
  year = {2024}
}

@inproceedings{bisk2020,
  title = {{PIQA}: Reasoning about Physical Commonsense in Natural Language},
  author = {Bisk, Yonatan and Zellers, Rowan and Le Bras, Ronan and Gao, Jianfeng and Choi, Yejin},
  booktitle = {AAAI Conference on Artificial Intelligence (AAAI)},
  year = {2020}
}

@article{clark2018,
  title = {Think you have Solved Question Answering? Try {ARC}, the {AI2} Reasoning Challenge},
  author = {Peter Clark and Isaac Cowhey and Oren Etzioni and Tushar Khot and Ashish Sabharwal and Carissa Schoenick and Oyvind Tafjord},
  journal = {arXiv preprint arXiv:1803.05457},
  year = {2018}
}

@inproceedings{hendrycks2021,
  title = {Measuring Massive Multitask Language Understanding},
  author = {Dan Hendrycks and Collin Burns and Steven Basart and Andy Zou and Mantas Mazeika and Dawn Song and Jacob Steinhardt},
  booktitle = {International Conference on Learning Representations (ICLR)},
  year = {2021}
}

@inproceedings{liu2020,
  title = {{LogiQA}: A Challenge Dataset for Machine Reading Comprehension with Logical Reasoning},
  author = {Liu, Jian and Cui, Leyang and Liu, Hanmeng and Huang, Dandan and Wang, Yile and Zhang, Yue},
  booktitle = {International Joint Conference on Artificial Intelligence (IJCAI)},
  year = {2020}
}

@inproceedings{mihaylov2018,
  title = {Can a Suit of Armor Conduct Electricity? A New Dataset for Open Book Question Answering},
  author = {Mihaylov, Todor and Clark, Peter and Khot, Tushar and Sabharwal, Ashish},
  booktitle = {Conference on Empirical Methods in Natural Language Processing (EMNLP)},
  year = {2018}
}

@inproceedings{nie2020,
  title = {Adversarial {NLI}: A New Benchmark for Natural Language Understanding},
  author = {Nie, Yixin and Williams, Adina and Dinan, Emily and Bansal, Mohit and Weston, Jason and Kiela, Douwe},
  booktitle = {Annual Meeting of the Association for Computational Linguistics (ACL)},
  year = {2020}
}

@inproceedings{pilehvar2019,
  title = {{W}i{C}: the Word-in-Context Dataset for Evaluating Context-Sensitive Meaning Representations},
  author = {Pilehvar, Mohammad Taher and Camacho-Collados, Jose},
  booktitle = {Conference of the North American Chapter of the Association for Computational Linguistics: Human Language Technologies (NAACL-HLT)},
  year = {2019}
}

@inproceedings{roemmele2011,
  title = {Choice of Plausible Alternatives: An Evaluation of Commonsense Causal Reasoning},
  author = {Roemmele, Melissa and Bejan, Cosmin Adrian and Gordon, Andrew S.},
  booktitle = {AAAI Spring Symposium on Logical Formalizations of Commonsense Reasoning (Commonsense)},
  year = {2011}
}

@inproceedings{sakaguchi2020,
  title = {{WinoGrande}: An Adversarial {Winograd} Schema Challenge at Scale},
  author = {Sakaguchi, Keisuke and Le Bras, Ronan and Bhagavatula, Chandra and Choi, Yejin},
  booktitle = {AAAI Conference on Artificial Intelligence (AAAI)},
  year = {2020}
}

@inproceedings{sap2019,
  title = {Social {IQ}a: Commonsense Reasoning about Social Interactions},
  author = {Sap, Maarten and Rashkin, Hannah and Chen, Derek and Le Bras, Ronan and Choi, Yejin},
  booktitle = {Conference on Empirical Methods in Natural Language Processing and International Joint Conference on Natural Language Processing (EMNLP-IJCNLP)},
  year = {2019}
}

@inproceedings{welbl2017,
  title = {Crowdsourcing Multiple Choice Science Questions},
  author = {Welbl, Johannes and Liu, Nelson F. and Gardner, Matt},
  booktitle = {Workshop on Noisy User-generated Text (WNUT)},
  year = {2017}
}

@inproceedings{zellers2019,
  title = {{H}ella{S}wag: Can a Machine Really Finish Your Sentence?},
  author = {Zellers, Rowan and Holtzman, Ari and Bisk, Yonatan and Farhadi, Ali and Choi, Yejin},
  booktitle = {Annual Meeting of the Association for Computational Linguistics (ACL)},
  year = {2019}
}

@article{austin2021mbpp,
  title = {Program Synthesis with Large Language Models},
  author = {Jacob Austin and Augustus Odena and Maxwell Nye and Maarten Bosma and Henryk Michalewski and David Dohan and Ellen Jiang and Carrie Cai and Michael Terry and Quoc Le and Charles Sutton},
  journal = {arXiv preprint arXiv:2108.07732},
  year = {2021}
}

@article{biderman2024lmeval,
  title = {Lessons from the Trenches on Reproducible Evaluation of Language Models},
  author = {Stella Biderman and Hailey Schoelkopf and Lintang Sutawika and Leo Gao and Jonathan Tow and Baber Abbasi and Alham Fikri Aji and Pawan Sasanka Ammanamanchi and Sidney Black and Jordan Clive and Anthony DiPofi and Julen Etxaniz and Benjamin Fattori and Jessica Zosa Forde and Charles Foster and Jeffrey Hsu and Mimansa Jaiswal and Wilson Y. Lee and Haonan Li and Charles Lovering and Niklas Muennighoff and Ellie Pavlick and Jason Phang and Aviya Skowron and Samson Tan and Xiangru Tang and Kevin A. Wang and Genta Indra Winata and François Yvon and Andy Zou},
  journal = {arXiv preprint arXiv:2405.14782},
  year = {2024},
  url = {https://arxiv.org/abs/2405.14782v2}
}

@article{chen2021humaneval,
  title = {Evaluating Large Language Models Trained on Code},
  author = {Mark Chen and Jerry Tworek and Heewoo Jun and Qiming Yuan and Henrique Ponde de Oliveira Pinto and Jared Kaplan and Harri Edwards and Yuri Burda and Nicholas Joseph and Greg Brockman and Alex Ray and Raul Puri and Gretchen Krueger and Michael Petrov and Heidy Khlaaf and Girish Sastry and Pamela Mishkin and Brooke Chan and Scott Gray and Nick Ryder and Mikhail Pavlov and Alethea Power and Lukasz Kaiser and Mohammad Bavarian and Clemens Winter and Philippe Tillet and Felipe Petroski Such and Dave Cummings and Matthias Plappert and Fotios Chantzis and Elizabeth Barnes and Ariel Herbert-Voss and William Hebgen Guss and Alex Nichol and Alex Paino and Nikolas Tezak and Jie Tang and Igor Babuschkin and Suchir Balaji and Shantanu Jain and William Saunders and Christopher Hesse and Andrew N. Carr and Jan Leike and Josh Achiam and Vedant Misra and Evan Morikawa and Alec Radford and Matthew Knight and Miles Brundage and Mira Murati and Katie Mayer and Peter Welinder and Bob McGrew and Dario Amodei and Sam McCandlish and Ilya Sutskever and Wojciech Zaremba},
  journal = {arXiv preprint arXiv:2107.03374},
  year = {2021}
}

@inproceedings{hendrycks2021math,
  title = {Measuring Mathematical Problem Solving With the {MATH} Dataset},
  author = {Hendrycks, Dan and Burns, Collin and Kadavath, Saurav and Arora, Akul and Basart, Steven and Tang, Eric and Song, Dawn and Steinhardt, Jacob},
  booktitle = {Neural Information Processing Systems Track on Datasets and Benchmarks (NeurIPS Datasets and Benchmarks)},
  year = {2021}
}

@inproceedings{casanueva2020,
  title = {Efficient Intent Detection with Dual Sentence Encoders},
  author = {Casanueva, I{\~n}igo and Tem{\v{c}}inas, Tadas and Gerz, Daniela and Henderson, Matthew and Vuli{\'c}, Ivan},
  booktitle = {Workshop on Natural Language Processing for Conversational AI (NLP4ConvAI)},
  year = {2020}
}

@article{coucke2018,
  title = {{Snips} Voice Platform: an embedded Spoken Language Understanding system for private-by-design voice interfaces},
  author = {Alice Coucke and Alaa Saade and Adrien Ball and Théodore Bluche and Alexandre Caulier and David Leroy and Clément Doumouro and Thibault Gisselbrecht and Francesco Caltagirone and Thibaut Lavril and Maël Primet and Joseph Dureau},
  journal = {arXiv preprint arXiv:1805.10190},
  year = {2018}
}

@inproceedings{lu2021,
  title = {{CodeXGLUE}: A Machine Learning Benchmark Dataset for Code Understanding and Generation},
  author = {Lu, Shuai and Guo, Daya and Ren, Shuo and Huang, Junjie and Svyatkovskiy, Alexey and Blanco, Ambrosio and Clement, Colin and Drain, Dawn and Jiang, Daxin and Tang, Duyu and Li, Ge and Zhou, Lidong and Shou, Linjun and Zhou, Long and Tufano, Michele and Gong, Ming and Zhou, Ming and Duan, Nan and Sundaresan, Neel and Deng, Shao Kun and Fu, Shengyu and Liu, Shujie},
  booktitle = {Neural Information Processing Systems Track on Datasets and Benchmarks (NeurIPS Datasets and Benchmarks)},
  year = {2021}
}

@inproceedings{mathew2021,
  title = {{HateXplain}: A Benchmark Dataset for Explainable Hate Speech Detection},
  author = {Mathew, Binny and Saha, Punyajoy and Yimam, Seid Muhie and Biemann, Chris and Goyal, Pawan and Mukherjee, Animesh},
  booktitle = {AAAI Conference on Artificial Intelligence (AAAI)},
  year = {2021}
}

@inproceedings{tuggener2020,
  title = {{LEDGAR}: A Large-Scale Multi-label Corpus for Text Classification of Legal Provisions in Contracts},
  author = {Tuggener, Don and von D{\"a}niken, Pius and Peetz, Thomas and Cieliebak, Mark},
  booktitle = {Language Resources and Evaluation Conference (LREC)},
  year = {2020}
}

@inproceedings{pal2022medmcqa,
  title = {{MedMCQA}: A Large-scale Multi-Subject Multi-Choice Dataset for Medical domain Question Answering},
  author = {Pal, Ankit and Umapathi, Logesh Kumar and Sankarasubbu, Malaikannan},
  booktitle = {Conference on Health, Inference, and Learning (CHIL)},
  year = {2022}
}

@article{zhong2017seq2sql,
  title = {{Seq2SQL}: Generating Structured Queries from Natural Language using Reinforcement Learning},
  author = {Victor Zhong and Caiming Xiong and Richard Socher},
  journal = {arXiv preprint arXiv:1709.00103},
  year = {2017}
}

@inproceedings{fitzgerald2023massive,
  title = {{MASSIVE}: A 1{M}-Example Multilingual Natural Language Understanding Dataset with 51 Typologically-Diverse Languages},
  author = {FitzGerald, Jack and Hench, Christopher and Peris, Charith and Mackie, Scott and Rottmann, Kay and Sanchez, Ana and Nash, Aaron and Urbach, Liam and Kakarala, Vishesh and Singh, Richa and Ranganath, Swetha and Crist, Laurie and Britan, Misha and Leeuwis, Wouter and Tur, Gokhan and Natarajan, Prem},
  booktitle = {Annual Meeting of the Association for Computational Linguistics (ACL)},
  year = {2023}
}

@inproceedings{bastianelli2020slurp,
  title = {{SLURP}: A Spoken Language Understanding Resource Package},
  author = {Bastianelli, Emanuele and Vanzo, Andrea and Swietojanski, Pawel and Rieser, Verena},
  booktitle = {Conference on Empirical Methods in Natural Language Processing (EMNLP)},
  year = {2020}
}

@inproceedings{lai2017race,
  title = {{RACE}: Large-scale {R}e{A}ding Comprehension Dataset From Examinations},
  author = {Lai, Guokun and Xie, Qizhe and Liu, Hanxiao and Yang, Yiming and Hovy, Eduard},
  booktitle = {Conference on Empirical Methods in Natural Language Processing (EMNLP)},
  year = {2017}
}

@inproceedings{chalkidis2022lexglue,
  title = {{L}ex{GLUE}: A Benchmark Dataset for Legal Language Understanding in {E}nglish},
  author = {Chalkidis, Ilias and Jana, Abhik and Hartung, Dirk and Bommarito, Michael and Androutsopoulos, Ion and Katz, Daniel Martin and Aletras, Nikolaos},
  booktitle = {Annual Meeting of the Association for Computational Linguistics (ACL)},
  year = {2022}
}

@inproceedings{dua2019drop,
  title = {{DROP}: A Reading Comprehension Benchmark Requiring Discrete Reasoning Over Paragraphs},
  author = {Dua, Dheeru and Wang, Yizhong and Dasigi, Pradeep and Stanovsky, Gabriel and Singh, Sameer and Gardner, Matt},
  booktitle = {Conference of the North American Chapter of the Association for Computational Linguistics: Human Language Technologies (NAACL-HLT)},
  year = {2019}
}

@inproceedings{hendrickx2010semeval,
  title = {{S}em{E}val-2010 Task 8: Multi-Way Classification of Semantic Relations between Pairs of Nominals},
  author = {Hendrickx, Iris and Kim, Su Nam and Kozareva, Zornitsa and Nakov, Preslav and {\'O} S{\'e}aghdha, Diarmuid and Pad{\'o}, Sebastian and Pennacchiotti, Marco and Romano, Lorenza and Szpakowicz, Stan},
  booktitle = {International Workshop on Semantic Evaluation (SemEval)},
  year = {2010}
}

@inproceedings{howard2018ulmfit,
  title = {Universal Language Model Fine-tuning for Text Classification},
  author = {Howard, Jeremy and Ruder, Sebastian},
  booktitle = {Annual Meeting of the Association for Computational Linguistics (ACL)},
  year = {2018}
}

@inproceedings{gururangan2020dontstop,
  title = {Don{'}t Stop Pretraining: Adapt Language Models to Domains and Tasks},
  author = {Gururangan, Suchin and Marasovi{\'c}, Ana and Swayamdipta, Swabha and Lo, Kyle and Beltagy, Iz and Downey, Doug and Smith, Noah A.},
  booktitle = {Annual Meeting of the Association for Computational Linguistics (ACL)},
  year = {2020}
}

@inproceedings{wei2022flan,
  title = {Finetuned Language Models Are Zero-Shot Learners},
  author = {Wei, Jason and Bosma, Maarten and Zhao, Vincent Y. and Guu, Kelvin and Yu, Adams Wei and Lester, Brian and Du, Nan and Dai, Andrew M. and Le, Quoc V.},
  booktitle = {International Conference on Learning Representations (ICLR)},
  year = {2022}
}

@inproceedings{sanh2022t0,
  title = {Multitask Prompted Training Enables Zero-Shot Task Generalization},
  author = {Sanh, Victor and Webson, Albert and Raffel, Colin and Bach, Stephen H. and Sutawika, Lintang and Alyafeai, Zaid and Chaffin, Antoine and Stiegler, Arnaud and Le Scao, Teven and Raja, Arun and Dey, Manan and Bari, M Saiful and Xu, Canwen and Thakker, Urmish and Sharma, Shanya and Szczechla, Eliza and Kim, Taewoon and Chhablani, Gunjan and Nayak, Nihal V. and Datta, Debajyoti and Chang, Jonathan and Jiang, Mike Tian-Jian and Wang, Han and Manica, Matteo and Shen, Sheng and Yong, Zheng-Xin and Pandey, Harshit and McKenna, Michael and Bawden, Rachel and Wang, Thomas and Neeraj, Trishala and Rozen, Jos and Sharma, Abheesht and Santilli, Andrea and Fevry, Thibault and Fries, Jason Alan and Teehan, Ryan and Bers, Tali and Biderman, Stella and Gao, Leo and Wolf, Thomas and Rush, Alexander M.},
  booktitle = {International Conference on Learning Representations (ICLR)},
  year = {2022}
}

@inproceedings{zenke2017si,
  title = {Continual Learning Through Synaptic Intelligence},
  author = {Friedemann Zenke and Ben Poole and Surya Ganguli},
  booktitle = {International Conference on Machine Learning (ICML)},
  year = {2017}
}

@inproceedings{aljundi2018mas,
  title = {Memory Aware Synapses: Learning what (not) to forget},
  author = {Aljundi, Rahaf and Babiloni, Francesca and Elhoseiny, Mohamed and Rohrbach, Marcus and Tuytelaars, Tinne},
  booktitle = {European Conference on Computer Vision (ECCV)},
  year = {2018}
}

@inproceedings{lopezpaz2017gem,
  title = {Gradient Episodic Memory for Continual Learning},
  author = {Lopez-Paz, David and Ranzato, Marc'Aurelio},
  booktitle = {Advances in Neural Information Processing Systems (NeurIPS)},
  year = {2017}
}

@inproceedings{chaudhry2019agem,
  title = {Efficient Lifelong Learning with {A-GEM}},
  author = {Chaudhry, Arslan and Ranzato, Marc'Aurelio and Rohrbach, Marcus and Elhoseiny, Mohamed},
  booktitle = {International Conference on Learning Representations (ICLR)},
  year = {2019}
}

@inproceedings{shin2017dgr,
  title = {Continual Learning with Deep Generative Replay},
  author = {Shin, Hanul and Lee, Jung Kwon and Kim, Jaehong and Kim, Jiwon},
  booktitle = {Advances in Neural Information Processing Systems (NeurIPS)},
  year = {2017}
}

@inproceedings{buzzega2020der,
  title = {Dark Experience for General Continual Learning: a Strong, Simple Baseline},
  author = {Buzzega, Pietro and Boschini, Matteo and Porrello, Angelo and Abati, Davide and Calderara, Simone},
  booktitle = {Advances in Neural Information Processing Systems (NeurIPS)},
  year = {2020}
}

@inproceedings{matena2022fisher,
  title = {Merging Models with {Fisher}-Weighted Averaging},
  author = {Matena, Michael S and Raffel, Colin},
  booktitle = {Advances in Neural Information Processing Systems (NeurIPS)},
  year = {2022}
}

@inproceedings{jin2023regmean,
  title = {Dataless Knowledge Fusion by Merging Weights of Language Models},
  author = {Jin, Xisen and Ren, Xiang and Preotiuc-Pietro, Daniel and Cheng, Pengxiang},
  booktitle = {International Conference on Learning Representations (ICLR)},
  year = {2023}
}

@inproceedings{yang2024adamerging,
  title = {{AdaMerging}: Adaptive Model Merging for Multi-Task Learning},
  author = {Yang, Enneng and Wang, Zhenyi and Shen, Li and Liu, Shiwei and Guo, Guibing and Wang, Xingwei and Tao, Dacheng},
  booktitle = {International Conference on Learning Representations (ICLR)},
  year = {2024}
}

@incollection{mccloskey1989catastrophic,
  title = {Catastrophic Interference in Connectionist Networks: The Sequential Learning Problem},
  author = {McCloskey, Michael and Cohen, Neal J.},
  booktitle = {Psychology of Learning and Motivation},
  volume = {24},
  pages = {109--165},
  year = {1989},
  publisher = {Elsevier},
  doi = {10.1016/s0079-7421(08)60536-8}
}

@inproceedings{yang2023neural,
  title = {Neural Collapse Inspired Feature-Classifier Alignment for Few-Shot Class Incremental Learning},
  author = {Yang, Yibo and Yuan, Haobo and Li, Xiangtai and Lin, Zhouchen and Torr, Philip and Tao, Dacheng},
  booktitle = {International Conference on Learning Representations (ICLR)},
  year = {2023}
}

@inproceedings{liu2025enhancing,
  title = {Enhancing Online Continual Learning with Plug-and-Play State Space Model and Class-Conditional Mixture of Discretization},
  author = {Liu, Sihao and Yang, Yibo and Li, Xiaojie and Clifton, David A. and Ghanem, Bernard},
  booktitle = {IEEE/CVF Conference on Computer Vision and Pattern Recognition (CVPR)},
  year = {2025}
}

@inproceedings{wang2024twostage,
  title = {Two-stage {LLM} Fine-tuning with Less Specialization and More Generalization},
  author = {Wang, Yihan and Si, Si and Li, Daliang and Lukasik, Michal and Yu, Felix and Hsieh, Cho-Jui and Dhillon, Inderjit and Kumar, Sanjiv},
  booktitle = {International Conference on Learning Representations (ICLR)},
  year = {2024}
}

@article{luo2025empirical,
  title = {An Empirical Study of Catastrophic Forgetting in Large Language Models During Continual Fine-Tuning},
  author = {Luo, Yun and Yang, Zhen and Meng, Fandong and Li, Yafu and Zhou, Jie and Zhang, Yue},
  journal = {IEEE Transactions on Audio, Speech and Language Processing},
  volume = {33},
  pages = {3776--3786},
  year = {2025},
  doi = {10.1109/taslpro.2025.3606231}
}

@article{kalajdzievski2024scaling,
  title = {Scaling Laws for Forgetting When Fine-Tuning Large Language Models},
  author = {Kalajdzievski, Damjan},
  journal = {arXiv preprint arXiv:2401.05605},
  year = {2024}
}

@article{ahmad2025opencodeinstruct,
  title = {{OpenCodeInstruct}: A Large-scale Instruction Tuning Dataset for Code {LLMs}},
  author = {Ahmad, Wasi Uddin and Ficek, Aleksander and Samadi, Mehrzad and Huang, Jocelyn and Noroozi, Vahid and Majumdar, Somshubra and Ginsburg, Boris},
  journal = {arXiv preprint arXiv:2504.04030},
  year = {2025}
}
\bibliographystyle{iclr2027_conference}

\clearpage
\appendix

\addtocontents{toc}{\protect\setcounter{tocdepth}{2}}
\begingroup
\renewcommand{\contentsname}{Appendix Contents}
\tableofcontents
\endgroup
\clearpage

\section{Implementation Details}
\label{app:implementation}
\label{app:experimental-details}
\label{app:training-settings}
\label{sec:training-settings}

\begin{table}[H]
\centering
\caption{Architecture, training, and inference settings of \name{}.}
\label{tab:implementation-details}
\small
\setlength{\tabcolsep}{5pt}
\renewcommand{\arraystretch}{1.02}
\begin{tabularx}{\linewidth}{@{}p{0.39\linewidth}X@{}}
\toprule
\textbf{Parameter} & \textbf{Value} \\
\midrule
\multicolumn{2}{@{}l}{\textbf{Repairing Side Net (\repairman{})}} \\
Hidden dimension / layers & 192 / 6 \\
Attention heads / feedforward dimension & 4 / 768 \\
Backbone layers (Qwen / Llama) & 5, 10, 15, 19, 24, 28 / 6, 11, 17, 22, 27, 32 \\
Prompt gate & Mean pooling, MLP (hidden dimension 384), sigmoid \\
Ladder gate & Learned per layer; sigmoid temperature 0.1 \\
Parameters (Qwen / Llama) & 9.57M / 10.56M (approximately 0.13\% of backbone) \\
\midrule
\multicolumn{2}{@{}l}{\textbf{Hypernetwork}} \\
Encoder hidden dimension / layers & 128 / 2 \\
Attention heads / feedforward dimension & 4 / 512 \\
Local / global encoding dimension & 128 / 256 \\
Maximum matrix-update rank & 8 \\
Fingerprint & Layerwise activation change: direction and log magnitude \\
Fingerprint prompts & 150 unlabeled general prompts; 10 per task \\
Shared template initialization & Mean of the source \repairman{} parameters \\
Generated update initialization & Zero \\
\midrule
\multicolumn{2}{@{}l}{\textbf{Domain LoRA fine-tuning}} \\
Rank / scaling parameter $\alpha$ & 64 / 128 \\
Target modules & Query, key, value, and output attention projections; gate, up, and down MLP projections \\
Epochs / effective batch size & 2 / 16 \\
Optimizer / learning rate & AdamW / $3\times10^{-4}$ \\
Scheduler / weight decay & Linear decay / 0 \\
Dropout / gradient clipping / seed & 0.05 / 1 / 42 \\
\midrule
\multicolumn{2}{@{}l}{\textbf{Ground-truth \repairman{} training}} \\
Training steps & 1,500 per source model \\
Optimizer / learning rate & AdamW / $10^{-4}$ \\
Learning-rate schedule & 200-step linear warmup, then constant \\
General examples per step & 64 labeled examples from general datasets \\
Domain prompts per step & 64 prompts without answer labels \\
\midrule
\multicolumn{2}{@{}l}{\textbf{Hypernetwork training}} \\
Training steps & 3,000 per backbone and split \\
Optimizer / learning rate & AdamW / $3\times10^{-5}$ (constant) \\
General examples per step & 64 labeled examples from general datasets \\
Domain prompts per step & 64 source-domain prompts without answer labels \\
Loss weights $(\beta_{\mathrm{func}},\beta_{\mathrm{weight}},\beta_{\mathrm{align}})$ & $(1,\,0.2,\,0.1)$ \\
Source / held-out domains & 9 / 3 per split \\
\midrule
\multicolumn{2}{@{}l}{\textbf{Inference}} \\
Repair generation & One \hypernet{} forward pass per model fingerprint \\
Total deployment time & $<1$ min on one NVIDIA A100 (40\,GB), including fingerprint extraction \\
Prompt-gate threshold & Values below 0.1 are set to zero \\
Parameter reuse & The generated \repairman{} is reused across requests \\
\bottomrule
\end{tabularx}
\end{table}

The source \repairman{}s and \hypernet{} use the same general training data (Table~\ref{tab:general-datasets}). We evaluate their final training checkpoints.

\paragraph{Parameter generation.}
\label{app:repairman-architecture}
\label{app:parameter-generation}
The fingerprint encoder adds projected direction and magnitude features to a learned layer embedding. Mean pooling of its layer encodings, followed by a nonlinear projection, gives the global encoding. Local parameter heads are shared across side layers; the input-embedding projection uses the first decoder layer's encoding together with the global encoding. Matrix heads predict low-rank factors, while vector and scalar heads predict their updates directly. Initializing one factor of each matrix update and all vector and scalar updates to zero makes the initial generated \repairman{} equal to the shared template. Activation means and template initialization use FP32, with $\epsilon=10^{-8}$ in Equation~\ref{eq:fingerprint}.

\paragraph{Functional supervision and answer scoring.}
\label{app:functional-objective}
For discriminative tasks, candidate-answer scores sum continuation-token log-probabilities. For tasks using length-normalized accuracy, training divides each score by the candidate's character count and multiplies it by 20 before applying option cross-entropy. For generative tasks, training uses the mean cross-entropy over response tokens under teacher forcing, with prompt and formatting tokens masked. The general-task loss is averaged over general examples. The positive and negative gate losses are averaged separately over general and domain prompts, respectively; these three means are added with unit coefficients. Gate supervision uses logits for numerical stability, and training uses the unthresholded gate.

\paragraph{Reconstruction and representation alignment.}
\label{app:hypernet-objective}
\label{app:representation-alignment}
We average the reconstruction MSE equally across parameter tensors and source domains. The error is normalized by its value at \hypernet{} initialization, computed with dropout disabled and kept fixed throughout training. Representation alignment compares the mean correction states over non-padding prompt tokens, before gating. It applies layer normalization without affine parameters ($\epsilon=10^{-5}$) to each pooled vector and averages squared differences over hidden dimensions and an equal mixture of general and source-domain prompts. The ground-truth \repairman{} remains frozen.

\section{Dataset Details}
\label{app:dataset-details}
\label{app:data-evaluation}

The domain datasets cover classification, question answering, semantic parsing, and code summarization (Table~\ref{tab:domain-datasets}). Domain answers are used for fine-tuning the backbone; repair training requires domain prompts without answer labels. General capabilities are evaluated on the 15 benchmarks in Table~\ref{tab:general-datasets}, which reports the shared general training pools for source \repairman{}s and \hypernet{}, along with the test set sizes used in this study. When a labeled official test set is unavailable, we use the official validation split for testing.

\paragraph{Evaluation.}
\label{app:general-evaluation}
General evaluation follows the prompts, few-shot settings, filters, and metrics of \texttt{lm-evaluation-harness} v0.4.12 \citep{biderman2024lmeval}, without a chat template. The average general score is the unweighted mean of the 15 task scores, computed before rounding. All scores are reported on a 0--100 scale.

\begin{table}[htbp]
\centering
\caption{The 12 adaptation datasets.}
\label{tab:domain-datasets}
\small
\setlength{\tabcolsep}{3pt}
\renewcommand{\arraystretch}{1.06}
\begin{tabularx}{\linewidth}{@{}>{\raggedright\arraybackslash}p{0.34\linewidth}>{\raggedright\arraybackslash}Xrr>{\raggedright\arraybackslash}p{0.18\linewidth}@{}}
\toprule
\textbf{Dataset} & \textbf{Task} & \textbf{Train} & \textbf{Test} & \textbf{Metric} \\
\midrule
HateXplain \citep{mathew2021} & Hate-speech detection & 4,000 & 1,000 & Macro-F1 \\
SemEval-2010 T8 \citep{hendrickx2010semeval} & Relation classification & 4,000 & 2,717 & Macro-F1 \\
SCOTUS \citep{chalkidis2022lexglue} & Legal issue classification & 5,000 & 1,400 & Macro-F1 \\
MedMCQA \citep{pal2022medmcqa} & Medical QA & 4,000 & 1,000 & Accuracy \\
MASSIVE-en \citep{fitzgerald2023massive,bastianelli2020slurp} & Intent recognition & 4,000 & 1,000 & Accuracy \\
WikiSQL \citep{zhong2017seq2sql} & Text-to-SQL & 4,000 & 1,000 & Execution accuracy \\
RACE \citep{lai2017race} & Reading comprehension & 4,000 & 1,000 & Accuracy \\
Banking77 \citep{casanueva2020} & Banking intent & 4,000 & 1,000 & Accuracy \\
CodeXGLUE-JS \citep{lu2021} & Code summarization & 4,000 & 1,000 & BLEU-4 \\
LEDGAR-Top10 \citep{tuggener2020} & Contract classification & 4,000 & 1,000 & Macro-F1 \\
SNIPS-7 \citep{coucke2018} & Assistant intent & 4,000 & 695 & Accuracy \\
DROP \citep{dua2019drop} & Discrete-reasoning QA & 4,000 & 1,000 & F1 \\
\bottomrule
\end{tabularx}
\vspace{3pt}
\begin{minipage}{\linewidth}
\footnotesize
\textit{Note.} LEDGAR-Top10 selects ten classes from the 100-class, single-label LEDGAR version in LexGLUE \citep{chalkidis2022lexglue}.
\end{minipage}
\end{table}

\begin{table}[!ht]
\centering
\caption{The 15 general benchmarks.}
\label{tab:general-datasets}
\small
\setlength{\tabcolsep}{5pt}
\renewcommand{\arraystretch}{1.06}
\begin{tabularx}{\linewidth}{@{}>{\raggedright\arraybackslash}p{0.43\linewidth}>{\raggedright\arraybackslash}Xrr@{}}
\toprule
\textbf{Dataset} & \textbf{Category} & \textbf{Train} & \textbf{Test} \\
\midrule
\multicolumn{4}{@{}l}{\textit{Discriminative tasks}} \\
\addlinespace[2pt]
ARC-Challenge \citep{clark2018} & Science QA & 889 & 1,172 \\
HellaSwag \citep{zellers2019} & Commonsense reasoning & 1,970 & 10,042 \\
MMLU \citep{hendrycks2021} & Multitask knowledge & 21,752 & 14,042 \\
OpenBookQA \citep{mihaylov2018} & Science QA & 1,970 & 500 \\
PIQA \citep{bisk2020} & Physical reasoning & 1,970 & 1,838 \\
SciQ \citep{welbl2017} & Science QA & 11,448 & 1,000 \\
WinoGrande \citep{sakaguchi2020} & Commonsense reasoning & 1,970 & 1,267 \\
ANLI-R1 \citep{nie2020} & Natural language inference & 1,970 & 1,000 \\
COPA \citep{roemmele2011} & Causal reasoning & 290 & 100 \\
SocialIQA \citep{sap2019} & Social reasoning & 1,970 & 1,954 \\
LogiQA \citep{liu2020} & Logical reasoning & 1,970 & 651 \\
WiC \citep{pilehvar2019} & Word sense disambiguation & 1,970 & 638 \\
\midrule
\multicolumn{4}{@{}l}{\textit{Generative tasks}} \\
\addlinespace[2pt]
MATH \citep{hendrycks2021math} & Mathematical reasoning & 7,466 & 5,000 \\
HumanEval \citep{chen2021humaneval} & Code generation & 7,788 & 164 \\
MBPP \citep{austin2021mbpp} & Code generation & 6,004 & 500 \\
\bottomrule
\end{tabularx}
\vspace{3pt}
\begin{minipage}{\linewidth}
\footnotesize
\textit{Note.} HumanEval uses a cleaned OpenCodeInstruct \citep{ahmad2025opencodeinstruct} proxy training pool; MBPP combines official training examples with OpenCodeInstruct augmentation.
\end{minipage}
\end{table}

\section{Splits}
\label{app:domain-splits}

Each split holds out three adaptation domains and trains \hypernet{} on models from the remaining nine (Table~\ref{tab:domain-splits}). Both backbones use the same partition, and every domain is held out once. Split D defines the main comparison in Table~\ref{tab:general-results}.

\begin{table}[!ht]
\centering
\caption{Held-out adaptation domains in each split.}
\label{tab:domain-splits}
\small
\setlength{\tabcolsep}{8pt}
\renewcommand{\arraystretch}{1.12}
\begin{tabular*}{\linewidth}{@{\extracolsep{\fill}}llll@{}}
\toprule
\textbf{Split} & \textbf{Domain 1} & \textbf{Domain 2} & \textbf{Domain 3} \\
\midrule
A & HateXplain & SemEval-2010 T8 & SCOTUS \\
B & MedMCQA & MASSIVE-en & WikiSQL \\
C & RACE & Banking77 & CodeXGLUE-JS \\
D & LEDGAR-Top10 & SNIPS-7 & DROP \\
\bottomrule
\end{tabular*}
\end{table}

\section{Fingerprint Sensitivity}
\label{app:fingerprint-prompts}
\label{sec:fingerprint-sensitivity}

The prompt-substitution experiment tests how the generated repair changes when its fingerprint is estimated from different general prompts. We construct three disjoint groups, $\mathcal{C}^{(0)}$, $\mathcal{C}^{(1)}$, and $\mathcal{C}^{(2)}$, denoted A, B, and C in Section~\ref{sec:ablation-fingerprint-sampling}. Each contains ten unlabeled prompts from each of the 15 general datasets, giving 150 prompts per group. The same groups are shared across adaptation domains and splits; all 450 prompts are excluded from \hypernet{} functional supervision.

\hypernet{} training and the primary evaluation use group A. At evaluation, we replace it with B or C and recompute the fingerprint, keeping the fine-tuned model and trained \hypernet{} fixed. These experiments use Qwen2.5-7B on the three held-out domains of Split D, evaluated with SGLang on the complete test sets used for that split.

\clearpage
\section{Qualitative Recovery Examples}
\label{app:qualitative-recovery}

\begingroup
\setlength{\tabcolsep}{5pt}
\renewcommand{\arraystretch}{1.08}
\small
\begin{longtable}{@{}>{\raggedright\arraybackslash}p{0.314\linewidth}>{\raggedright\arraybackslash}p{0.314\linewidth}>{\raggedright\arraybackslash}p{0.314\linewidth}@{}}
\caption{Recovery examples on 13 additional benchmarks, showing abbreviated prompts, selected answers, and verbatim response excerpts.}\label{tab:recovery-cases}\\
\toprule
\textbf{Base (correct)} & \textbf{Fine-tuned (incorrect)} & \textbf{HyperRepair (correct)} \\
\midrule
\endfirsthead
\multicolumn{3}{c}{\textbf{Table \thetable{} (continued)}}\\[4pt]
\toprule
\textbf{Base (correct)} & \textbf{Fine-tuned (incorrect)} & \textbf{HyperRepair (correct)} \\
\midrule
\endhead
\bottomrule
\endfoot
\multicolumn{3}{@{}p{\linewidth}@{}}{\textbf{HellaSwag}\hfill\textit{Adaptation: CodeXGLUE-JS (Split C)}}\\*
\multicolumn{3}{@{}p{\linewidth}@{}}{\textbf{Prompt.} Clean and jerk: A woman in a black shirt is bent over. She}\\*[5pt]
picks up a heavy weight. & is putting lotion onto her back. & picks up a heavy weight. \\
\addlinespace[8pt]\midrule
\multicolumn{3}{@{}p{\linewidth}@{}}{\textbf{MMLU}\hfill\textit{Adaptation: SCOTUS (Split A)}}\\*
\multicolumn{3}{@{}p{\linewidth}@{}}{\textbf{Prompt.} A wave transfers}\\*[5pt]
energy & amplitude & energy \\
\addlinespace[8pt]\midrule
\multicolumn{3}{@{}p{\linewidth}@{}}{\textbf{OpenBookQA}\hfill\textit{Adaptation: MASSIVE-en (Split B)}}\\*
\multicolumn{3}{@{}p{\linewidth}@{}}{\textbf{Prompt.} Which musical instrument is the same type as a guitar?}\\*[5pt]
cello & drum & cello \\
\addlinespace[8pt]\midrule
\multicolumn{3}{@{}p{\linewidth}@{}}{\textbf{PIQA}\hfill\textit{Adaptation: SemEval-2010 T8 (Split A)}}\\*
\multicolumn{3}{@{}p{\linewidth}@{}}{\textbf{Prompt.} To place a screw in a plastic lid, you can}\\*[5pt]
Use a drill & Use a toothbrush & Use a drill \\
\addlinespace[8pt]\midrule
\multicolumn{3}{@{}p{\linewidth}@{}}{\textbf{SciQ}\hfill\textit{Adaptation: HateXplain (Split A)}}\\*
\multicolumn{3}{@{}p{\linewidth}@{}}{\textbf{Prompt.} What is it called when ice changes to liquid water?}\\*[5pt]
melting & condensing & melting \\
\addlinespace[8pt]\midrule
\multicolumn{3}{@{}p{\linewidth}@{}}{\textbf{WinoGrande}\hfill\textit{Adaptation: Banking77 (Split C)}}\\*
\multicolumn{3}{@{}p{\linewidth}@{}}{\textbf{Prompt.} Adam loved dogs but Jason was afraid of them, so only \_ petted the poodle.}\\*[5pt]
Adam & Jason & Adam \\
\addlinespace[8pt]\midrule
\multicolumn{3}{@{}p{\linewidth}@{}}{\textbf{ANLI-R1}\hfill\textit{Adaptation: DROP (Split D)}}\\*
\multicolumn{3}{@{}p{\linewidth}@{}}{\textbf{Prompt.} "Can't Help Thinking About Me" is a song written by David Bowie in 1965 and released as a single under the name David Bowie with The Lower Third. This was the first single released after he changed his name from David (also Davie) Jones to David Bowie.\par Question: David Bowie wrote the song can't help thinking about me in 1966. True, False, or Neither?}\\*[5pt]
False & True & False \\
\addlinespace[8pt]\midrule
\multicolumn{3}{@{}p{\linewidth}@{}}{\textbf{COPA}\hfill\textit{Adaptation: SNIPS-7 (Split D)}}\\*
\multicolumn{3}{@{}p{\linewidth}@{}}{\textbf{Prompt.} Everyone in the class turned to stare at the student.\par What was the cause?}\\*[5pt]
The student's phone rang. & The student took notes. & The student's phone rang. \\
\addlinespace[8pt]\midrule
\multicolumn{3}{@{}p{\linewidth}@{}}{\textbf{SocialIQA}\hfill\textit{Adaptation: HateXplain (Split A)}}\\*
\multicolumn{3}{@{}p{\linewidth}@{}}{\textbf{Prompt.} Jordan won a car and could not believe his luck.\par How would Jordan feel afterwards?}\\*[5pt]
thrilled & angry & thrilled \\
\addlinespace[8pt]\midrule
\multicolumn{3}{@{}p{\linewidth}@{}}{\textbf{LogiQA}\hfill\textit{Adaptation: MASSIVE-en (Split B)}}\\*
\multicolumn{3}{@{}p{\linewidth}@{}}{\textbf{Prompt.} Every politician has to please his voters. Although Ma Ying-jeou is an honest person, he is also a politician. No one can please his voters if he does not occasionally say something ambiguous.\par If the above statement is true, which of the following statements must be true?}\\*[5pt]
Ma Ying-jeou will say something ambiguous & The politicians who say ambiguously are not honest people. & Ma Ying-jeou will say something ambiguous \\
\addlinespace[8pt]\midrule
\multicolumn{3}{@{}p{\linewidth}@{}}{\textbf{WiC}\hfill\textit{Adaptation: SCOTUS (Split A)}}\\*
\multicolumn{3}{@{}p{\linewidth}@{}}{\textbf{Prompt.} Sentence 1: Burn a CD.\par Sentence 2: My eyes are burning.\par Question: Is the word 'Burn' used in the same way in the two sentences above?}\\*[5pt]
no & yes & no \\
\addlinespace[8pt]\midrule
\multicolumn{3}{@{}p{\linewidth}@{}}{\textbf{HumanEval}\hfill\textit{Adaptation: CodeXGLUE-JS (Split C)}}\\*
\multicolumn{3}{@{}p{\linewidth}@{}}{\textbf{Prompt.} def add(lst): Given a non-empty list of integers lst. add the even elements that are at odd indices..\par Example: add([4, 2, 6, 7]) ==> 2}\\*[5pt]
{\ttfamily\footnotesize return sum([lst[i] for i in range(1, len(lst), 2) if lst[i] \% 2 == 0])} & {\ttfamily\footnotesize return sum([lst[i] for i in range(len(lst)) if i \% 2 == 0 and lst[i] \% 2 == 0])} & {\ttfamily\footnotesize return sum(lst[i] for i in range(1, len(lst), 2) if lst[i] \% 2 == 0)} \\
\addlinespace[8pt]\midrule
\multicolumn{3}{@{}p{\linewidth}@{}}{\textbf{MBPP}\hfill\textit{Adaptation: WikiSQL (Split B)}}\\*
\multicolumn{3}{@{}p{\linewidth}@{}}{\textbf{Prompt.} Write a function to add the given list to the given tuples.\par Provided test: assert add\_lists([5, 6, 7], (9, 10)) == (9, 10, 5, 6, 7)}\\*[5pt]
{\ttfamily\footnotesize def add\_lists(list1, tuple1):\par \hspace*{0.9em}return tuple1 + tuple(list1)} & {\ttfamily\footnotesize def add\_lists(nums, tup):\par \hspace*{0.9em}return tup + nums} & {\ttfamily\footnotesize def add\_lists(test\_list, test\_tuple):\par \hspace*{0.9em}res = test\_tuple + tuple(test\_list)\par \hspace*{0.9em}return (res)} \\
\end{longtable}
\endgroup

\clearpage
\section{Additional Split Results}
\label{app:cross-split-evaluation}

In the following result tables, each entry shows Fine-tuned $\rightarrow$ \name{}.

\subsection{Split A: Qwen2.5-7B}
\begingroup
\newcommand{\scorepair}[2]{\mbox{#1\(\scriptstyle\rightarrow\)#2}}
\begin{center}
\captionsetup{type=table,font=small,skip=5pt}
\caption{Qwen2.5-7B results on Split A.}
\label{tab:split-qwen-A}
\fontsize{8}{9}\selectfont
\setlength{\tabcolsep}{1pt}
\renewcommand{\arraystretch}{1.0}
\begin{tabular*}{\linewidth}{@{\extracolsep{\fill}}ll*{6}{c}@{}}
\toprule
\textbf{Role} & \textbf{Domain} & \textbf{Domain score} & \textbf{ARC-C} & \textbf{HellaSwag} & \textbf{MMLU} & \textbf{OpenBookQA} & \textbf{PIQA} \\
\midrule
Held-out & HateXplain & \scorepair{68.3}{67.6} & \scorepair{41.1}{56.9} & \scorepair{63.0}{83.8} & \scorepair{67.4}{65.7} & \scorepair{43.2}{50.6} & \scorepair{73.3}{80.6} \\
 & SemEval-2010 T8 & \scorepair{87.5}{87.5} & \scorepair{51.4}{60.2} & \scorepair{72.9}{86.2} & \scorepair{63.3}{62.6} & \scorepair{46.6}{50.0} & \scorepair{77.3}{82.6} \\
 & SCOTUS & \scorepair{68.3}{67.7} & \scorepair{45.7}{61.3} & \scorepair{77.7}{89.0} & \scorepair{25.6}{37.9} & \scorepair{46.8}{53.0} & \scorepair{80.2}{84.1} \\
\midrule
Source & MedMCQA & \scorepair{38.4}{38.1} & \scorepair{62.5}{70.9} & \scorepair{79.3}{90.5} & \scorepair{69.5}{70.0} & \scorepair{50.2}{57.6} & \scorepair{81.4}{84.7} \\
 & MASSIVE-en & \scorepair{85.7}{85.7} & \scorepair{56.7}{69.4} & \scorepair{78.1}{90.0} & \scorepair{68.9}{69.5} & \scorepair{47.8}{57.2} & \scorepair{79.7}{84.8} \\
 & WikiSQL & \scorepair{86.5}{86.3} & \scorepair{55.9}{65.8} & \scorepair{77.3}{90.2} & \scorepair{66.6}{68.0} & \scorepair{44.4}{52.6} & \scorepair{79.4}{84.7} \\
\addlinespace[3pt]
 & RACE & \scorepair{59.5}{59.6} & \scorepair{60.4}{70.1} & \scorepair{79.5}{90.1} & \scorepair{68.9}{70.3} & \scorepair{48.0}{56.6} & \scorepair{79.9}{85.2} \\
 & Banking77 & \scorepair{92.7}{92.7} & \scorepair{53.4}{64.4} & \scorepair{76.7}{89.3} & \scorepair{27.3}{49.5} & \scorepair{45.4}{56.8} & \scorepair{77.4}{83.0} \\
 & CodeXGLUE-JS & \scorepair{15.8}{15.9} & \scorepair{55.7}{69.2} & \scorepair{79.2}{90.8} & \scorepair{71.3}{71.5} & \scorepair{45.6}{53.2} & \scorepair{78.8}{84.9} \\
\addlinespace[3pt]
 & LEDGAR-Top10 & \scorepair{98.0}{98.2} & \scorepair{53.8}{66.9} & \scorepair{73.1}{89.8} & \scorepair{55.0}{64.6} & \scorepair{46.0}{56.0} & \scorepair{79.4}{84.4} \\
 & SNIPS-7 & \scorepair{98.3}{98.3} & \scorepair{44.8}{63.3} & \scorepair{70.3}{88.1} & \scorepair{61.8}{64.9} & \scorepair{41.8}{52.8} & \scorepair{75.1}{82.8} \\
 & DROP & \scorepair{76.2}{75.9} & \scorepair{59.9}{71.0} & \scorepair{76.1}{89.3} & \scorepair{64.8}{66.1} & \scorepair{46.8}{55.0} & \scorepair{79.8}{84.8} \\
\bottomrule
\end{tabular*}
\par\vspace{6pt}
\begin{tabular*}{\linewidth}{@{\extracolsep{\fill}}ll*{5}{c}@{}}
\toprule
\textbf{Role} & \textbf{Domain} & \textbf{SciQ} & \textbf{WinoGrande} & \textbf{ANLI-R1} & \textbf{SocialIQA} & \textbf{LogiQA} \\
\midrule
Held-out & HateXplain & \scorepair{77.6}{83.3} & \scorepair{65.6}{76.2} & \scorepair{40.5}{46.8} & \scorepair{40.5}{42.3} & \scorepair{31.6}{32.9} \\
 & SemEval-2010 T8 & \scorepair{80.9}{85.8} & \scorepair{69.4}{78.5} & \scorepair{50.6}{51.9} & \scorepair{45.9}{47.5} & \scorepair{33.8}{32.0} \\
 & SCOTUS & \scorepair{85.0}{88.6} & \scorepair{73.6}{78.5} & \scorepair{40.4}{43.7} & \scorepair{47.6}{49.3} & \scorepair{31.8}{33.6} \\
\midrule
Source & MedMCQA & \scorepair{95.7}{97.6} & \scorepair{72.0}{79.2} & \scorepair{53.1}{59.7} & \scorepair{59.3}{60.3} & \scorepair{35.5}{46.7} \\
 & MASSIVE-en & \scorepair{96.0}{96.9} & \scorepair{74.2}{79.0} & \scorepair{44.7}{54.6} & \scorepair{50.0}{53.2} & \scorepair{34.9}{42.2} \\
 & WikiSQL & \scorepair{96.9}{97.7} & \scorepair{73.0}{80.7} & \scorepair{51.9}{60.9} & \scorepair{53.2}{57.1} & \scorepair{34.1}{38.2} \\
\addlinespace[3pt]
 & RACE & \scorepair{96.5}{97.4} & \scorepair{69.1}{78.5} & \scorepair{50.8}{65.5} & \scorepair{57.5}{60.3} & \scorepair{48.2}{50.7} \\
 & Banking77 & \scorepair{95.0}{96.2} & \scorepair{72.6}{80.7} & \scorepair{35.0}{45.9} & \scorepair{47.2}{48.4} & \scorepair{33.2}{34.6} \\
 & CodeXGLUE-JS & \scorepair{96.5}{97.3} & \scorepair{70.7}{79.2} & \scorepair{52.4}{61.0} & \scorepair{54.7}{57.2} & \scorepair{39.3}{45.9} \\
\addlinespace[3pt]
 & LEDGAR-Top10 & \scorepair{88.1}{94.5} & \scorepair{70.8}{78.2} & \scorepair{48.5}{53.6} & \scorepair{49.0}{52.0} & \scorepair{33.6}{33.8} \\
 & SNIPS-7 & \scorepair{87.3}{93.8} & \scorepair{68.7}{78.7} & \scorepair{37.1}{46.9} & \scorepair{38.0}{42.5} & \scorepair{31.0}{36.7} \\
 & DROP & \scorepair{97.3}{97.1} & \scorepair{75.1}{81.1} & \scorepair{47.0}{61.2} & \scorepair{54.9}{59.5} & \scorepair{44.9}{43.3} \\
\bottomrule
\end{tabular*}
\par\vspace{6pt}
\begin{tabular*}{\linewidth}{@{\extracolsep{\fill}}ll*{5}{c}@{}}
\toprule
\textbf{Role} & \textbf{Domain} & \textbf{WiC} & \textbf{COPA} & \textbf{MATH} & \textbf{HumanEval} & \textbf{MBPP} \\
\midrule
Held-out & HateXplain & \scorepair{50.0}{53.8} & \scorepair{75}{92} & \scorepair{0.0}{37.2} & \scorepair{0.0}{41.5} & \scorepair{0.0}{50.6} \\
 & SemEval-2010 T8 & \scorepair{49.7}{50.8} & \scorepair{85}{90} & \scorepair{0.0}{41.6} & \scorepair{0.0}{43.9} & \scorepair{6.4}{52.0} \\
 & SCOTUS & \scorepair{50.3}{58.5} & \scorepair{89}{92} & \scorepair{0.0}{43.9} & \scorepair{22.6}{32.9} & \scorepair{36.0}{29.0} \\
\midrule
Source & MedMCQA & \scorepair{55.8}{68.0} & \scorepair{90}{92} & \scorepair{40.2}{43.8} & \scorepair{56.1}{50.0} & \scorepair{62.8}{58.4} \\
 & MASSIVE-en & \scorepair{64.9}{64.7} & \scorepair{91}{95} & \scorepair{43.4}{45.6} & \scorepair{46.3}{43.3} & \scorepair{42.0}{46.0} \\
 & WikiSQL & \scorepair{60.5}{63.3} & \scorepair{91}{95} & \scorepair{39.7}{41.3} & \scorepair{52.4}{47.0} & \scorepair{59.4}{55.2} \\
\addlinespace[3pt]
 & RACE & \scorepair{51.6}{63.8} & \scorepair{84}{92} & \scorepair{34.7}{43.4} & \scorepair{54.3}{47.6} & \scorepair{62.4}{60.4} \\
 & Banking77 & \scorepair{56.6}{58.2} & \scorepair{90}{91} & \scorepair{15.0}{40.1} & \scorepair{43.3}{42.1} & \scorepair{47.6}{41.0} \\
 & CodeXGLUE-JS & \scorepair{56.0}{67.9} & \scorepair{89}{91} & \scorepair{22.0}{45.0} & \scorepair{53.7}{51.2} & \scorepair{60.4}{58.2} \\
\addlinespace[3pt]
 & LEDGAR-Top10 & \scorepair{50.0}{58.6} & \scorepair{88}{93} & \scorepair{3.6}{38.8} & \scorepair{22.0}{33.5} & \scorepair{28.8}{42.2} \\
 & SNIPS-7 & \scorepair{50.3}{54.7} & \scorepair{85}{93} & \scorepair{0.0}{35.4} & \scorepair{1.8}{34.1} & \scorepair{0.0}{28.2} \\
 & DROP & \scorepair{65.8}{65.4} & \scorepair{90}{96} & \scorepair{33.4}{39.7} & \scorepair{51.8}{47.0} & \scorepair{58.8}{56.6} \\
\bottomrule
\end{tabular*}
\end{center}
\endgroup

\clearpage
\subsection{Split A: Llama-3.1-8B}
\begingroup
\newcommand{\scorepair}[2]{\mbox{#1\(\scriptstyle\rightarrow\)#2}}
\begin{center}
\captionsetup{type=table,font=small,skip=5pt}
\caption{Llama-3.1-8B results on Split A.}
\label{tab:split-llama-A}
\fontsize{8}{9}\selectfont
\setlength{\tabcolsep}{1pt}
\renewcommand{\arraystretch}{1.0}
\begin{tabular*}{\linewidth}{@{\extracolsep{\fill}}ll*{6}{c}@{}}
\toprule
\textbf{Role} & \textbf{Domain} & \textbf{Domain score} & \textbf{ARC-C} & \textbf{HellaSwag} & \textbf{MMLU} & \textbf{OpenBookQA} & \textbf{PIQA} \\
\midrule
Held-out & HateXplain & \scorepair{65.6}{63.8} & \scorepair{35.5}{43.8} & \scorepair{51.6}{65.4} & \scorepair{36.8}{34.6} & \scorepair{39.0}{45.2} & \scorepair{67.5}{75.4} \\
 & SemEval-2010 T8 & \scorepair{89.0}{88.9} & \scorepair{42.7}{50.3} & \scorepair{73.9}{80.6} & \scorepair{51.6}{51.4} & \scorepair{43.8}{49.6} & \scorepair{78.1}{80.9} \\
 & SCOTUS & \scorepair{68.5}{68.5} & \scorepair{47.7}{52.8} & \scorepair{78.1}{82.5} & \scorepair{35.4}{41.6} & \scorepair{45.6}{49.2} & \scorepair{79.4}{82.0} \\
\midrule
Source & MedMCQA & \scorepair{40.0}{39.9} & \scorepair{56.5}{60.8} & \scorepair{79.8}{84.9} & \scorepair{45.3}{49.6} & \scorepair{49.0}{52.8} & \scorepair{81.5}{82.1} \\
 & MASSIVE-en & \scorepair{84.2}{84.2} & \scorepair{46.7}{55.1} & \scorepair{75.0}{82.7} & \scorepair{41.1}{37.2} & \scorepair{42.8}{51.0} & \scorepair{77.4}{81.9} \\
 & WikiSQL & \scorepair{85.6}{85.5} & \scorepair{49.5}{54.2} & \scorepair{76.6}{83.6} & \scorepair{52.0}{53.9} & \scorepair{42.4}{49.0} & \scorepair{79.4}{81.7} \\
\addlinespace[3pt]
 & RACE & \scorepair{59.6}{59.4} & \scorepair{58.8}{62.4} & \scorepair{78.9}{83.9} & \scorepair{47.6}{48.3} & \scorepair{47.0}{52.0} & \scorepair{81.5}{82.6} \\
 & Banking77 & \scorepair{92.9}{92.9} & \scorepair{49.2}{54.0} & \scorepair{77.0}{82.3} & \scorepair{35.0}{34.2} & \scorepair{47.0}{50.4} & \scorepair{78.0}{81.3} \\
 & CodeXGLUE-JS & \scorepair{15.6}{15.6} & \scorepair{53.4}{57.8} & \scorepair{78.8}{85.0} & \scorepair{62.7}{61.5} & \scorepair{46.0}{52.2} & \scorepair{80.6}{82.1} \\
\addlinespace[3pt]
 & LEDGAR-Top10 & \scorepair{98.0}{98.0} & \scorepair{40.1}{52.5} & \scorepair{62.6}{80.0} & \scorepair{28.1}{29.9} & \scorepair{42.8}{50.2} & \scorepair{74.0}{80.8} \\
 & SNIPS-7 & \scorepair{98.7}{98.7} & \scorepair{38.1}{49.7} & \scorepair{68.3}{78.1} & \scorepair{23.5}{25.8} & \scorepair{41.0}{47.6} & \scorepair{75.9}{80.4} \\
 & DROP & \scorepair{65.4}{65.3} & \scorepair{55.2}{59.6} & \scorepair{76.2}{82.1} & \scorepair{33.2}{36.7} & \scorepair{47.2}{52.0} & \scorepair{79.8}{82.1} \\
\bottomrule
\end{tabular*}
\par\vspace{6pt}
\begin{tabular*}{\linewidth}{@{\extracolsep{\fill}}ll*{5}{c}@{}}
\toprule
\textbf{Role} & \textbf{Domain} & \textbf{SciQ} & \textbf{WinoGrande} & \textbf{ANLI-R1} & \textbf{SocialIQA} & \textbf{LogiQA} \\
\midrule
Held-out & HateXplain & \scorepair{64.4}{73.0} & \scorepair{66.7}{73.0} & \scorepair{33.3}{33.3} & \scorepair{36.6}{38.2} & \scorepair{26.3}{24.3} \\
 & SemEval-2010 T8 & \scorepair{72.0}{78.7} & \scorepair{71.6}{78.2} & \scorepair{34.9}{40.2} & \scorepair{45.6}{45.9} & \scorepair{29.5}{26.6} \\
 & SCOTUS & \scorepair{80.1}{87.0} & \scorepair{72.2}{77.9} & \scorepair{34.2}{40.9} & \scorepair{47.1}{46.9} & \scorepair{32.0}{30.4} \\
\midrule
Source & MedMCQA & \scorepair{95.7}{96.4} & \scorepair{73.5}{79.2} & \scorepair{34.4}{44.1} & \scorepair{52.5}{53.6} & \scorepair{29.3}{28.6} \\
 & MASSIVE-en & \scorepair{84.4}{93.9} & \scorepair{74.0}{78.4} & \scorepair{32.3}{46.1} & \scorepair{42.2}{44.7} & \scorepair{30.7}{28.6} \\
 & WikiSQL & \scorepair{96.0}{96.5} & \scorepair{73.2}{80.2} & \scorepair{32.6}{38.6} & \scorepair{48.5}{50.8} & \scorepair{30.9}{30.7} \\
\addlinespace[3pt]
 & RACE & \scorepair{96.3}{97.2} & \scorepair{71.9}{78.9} & \scorepair{35.0}{43.1} & \scorepair{52.0}{54.3} & \scorepair{28.0}{28.3} \\
 & Banking77 & \scorepair{95.5}{94.5} & \scorepair{73.6}{78.5} & \scorepair{37.3}{47.0} & \scorepair{43.9}{46.5} & \scorepair{29.6}{28.4} \\
 & CodeXGLUE-JS & \scorepair{95.6}{97.2} & \scorepair{73.2}{79.0} & \scorepair{36.8}{44.3} & \scorepair{49.6}{50.8} & \scorepair{31.6}{33.0} \\
\addlinespace[3pt]
 & LEDGAR-Top10 & \scorepair{69.0}{91.9} & \scorepair{71.0}{77.4} & \scorepair{36.2}{42.5} & \scorepair{40.7}{44.0} & \scorepair{31.2}{28.4} \\
 & SNIPS-7 & \scorepair{76.7}{89.5} & \scorepair{70.7}{76.6} & \scorepair{33.2}{39.2} & \scorepair{40.3}{44.5} & \scorepair{29.3}{29.2} \\
 & DROP & \scorepair{96.3}{96.9} & \scorepair{74.5}{80.0} & \scorepair{32.1}{40.7} & \scorepair{51.5}{52.6} & \scorepair{27.5}{26.9} \\
\bottomrule
\end{tabular*}
\par\vspace{6pt}
\begin{tabular*}{\linewidth}{@{\extracolsep{\fill}}ll*{5}{c}@{}}
\toprule
\textbf{Role} & \textbf{Domain} & \textbf{WiC} & \textbf{COPA} & \textbf{MATH} & \textbf{HumanEval} & \textbf{MBPP} \\
\midrule
Held-out & HateXplain & \scorepair{50.0}{50.0} & \scorepair{71}{88} & \scorepair{0.0}{1.2} & \scorepair{0.0}{15.9} & \scorepair{0.0}{9.4} \\
 & SemEval-2010 T8 & \scorepair{50.0}{58.5} & \scorepair{89}{86} & \scorepair{0.0}{7.2} & \scorepair{0.6}{25.6} & \scorepair{0.0}{29.2} \\
 & SCOTUS & \scorepair{52.2}{50.2} & \scorepair{94}{93} & \scorepair{0.0}{9.3} & \scorepair{28.0}{25.6} & \scorepair{35.2}{29.8} \\
\midrule
Source & MedMCQA & \scorepair{59.9}{60.2} & \scorepair{90}{93} & \scorepair{12.2}{14.8} & \scorepair{34.8}{30.5} & \scorepair{43.2}{34.0} \\
 & MASSIVE-en & \scorepair{50.5}{54.7} & \scorepair{90}{91} & \scorepair{0.0}{9.7} & \scorepair{0.6}{21.3} & \scorepair{0.0}{13.0} \\
 & WikiSQL & \scorepair{50.9}{55.3} & \scorepair{87}{92} & \scorepair{6.1}{9.7} & \scorepair{20.1}{22.6} & \scorepair{28.6}{27.6} \\
\addlinespace[3pt]
 & RACE & \scorepair{50.2}{58.5} & \scorepair{91}{96} & \scorepair{3.3}{7.7} & \scorepair{34.1}{30.5} & \scorepair{41.6}{37.6} \\
 & Banking77 & \scorepair{49.8}{55.3} & \scorepair{91}{92} & \scorepair{0.0}{10.0} & \scorepair{25.6}{26.2} & \scorepair{29.0}{27.8} \\
 & CodeXGLUE-JS & \scorepair{58.3}{63.9} & \scorepair{93}{94} & \scorepair{5.9}{14.8} & \scorepair{33.5}{31.1} & \scorepair{42.2}{35.6} \\
\addlinespace[3pt]
 & LEDGAR-Top10 & \scorepair{50.0}{56.4} & \scorepair{86}{94} & \scorepair{0.0}{4.1} & \scorepair{0.0}{12.2} & \scorepair{0.0}{11.2} \\
 & SNIPS-7 & \scorepair{49.8}{54.1} & \scorepair{88}{90} & \scorepair{0.0}{3.9} & \scorepair{12.2}{14.6} & \scorepair{1.2}{10.4} \\
 & DROP & \scorepair{53.1}{60.5} & \scorepair{92}{94} & \scorepair{0.0}{3.4} & \scorepair{13.4}{22.0} & \scorepair{25.4}{33.8} \\
\bottomrule
\end{tabular*}
\end{center}
\endgroup

\clearpage
\subsection{Split B: Qwen2.5-7B}
\begingroup
\newcommand{\scorepair}[2]{\mbox{#1\(\scriptstyle\rightarrow\)#2}}
\begin{center}
\captionsetup{type=table,font=small,skip=5pt}
\caption{Qwen2.5-7B results on Split B.}
\label{tab:split-qwen-B}
\fontsize{8}{9}\selectfont
\setlength{\tabcolsep}{1pt}
\renewcommand{\arraystretch}{1.0}
\begin{tabular*}{\linewidth}{@{\extracolsep{\fill}}ll*{6}{c}@{}}
\toprule
\textbf{Role} & \textbf{Domain} & \textbf{Domain score} & \textbf{ARC-C} & \textbf{HellaSwag} & \textbf{MMLU} & \textbf{OpenBookQA} & \textbf{PIQA} \\
\midrule
Held-out & MedMCQA & \scorepair{38.4}{37.2} & \scorepair{62.5}{72.0} & \scorepair{79.3}{89.8} & \scorepair{69.5}{70.3} & \scorepair{50.2}{55.4} & \scorepair{81.4}{84.0} \\
 & MASSIVE-en & \scorepair{85.7}{85.7} & \scorepair{56.7}{69.5} & \scorepair{78.1}{88.7} & \scorepair{68.9}{69.4} & \scorepair{47.8}{55.2} & \scorepair{79.7}{83.9} \\
 & WikiSQL & \scorepair{86.5}{86.1} & \scorepair{55.9}{65.7} & \scorepair{77.3}{88.9} & \scorepair{66.6}{65.4} & \scorepair{44.4}{50.6} & \scorepair{79.4}{83.6} \\
\midrule
Source & HateXplain & \scorepair{68.3}{68.6} & \scorepair{41.1}{64.4} & \scorepair{63.0}{89.2} & \scorepair{67.4}{69.7} & \scorepair{43.2}{52.4} & \scorepair{73.3}{84.0} \\
 & SemEval-2010 T8 & \scorepair{87.5}{87.5} & \scorepair{51.4}{67.8} & \scorepair{72.9}{87.8} & \scorepair{63.3}{65.0} & \scorepair{46.6}{55.2} & \scorepair{77.3}{83.1} \\
 & SCOTUS & \scorepair{68.3}{68.4} & \scorepair{45.7}{65.0} & \scorepair{77.7}{89.5} & \scorepair{25.6}{62.8} & \scorepair{46.8}{54.0} & \scorepair{80.2}{83.9} \\
\addlinespace[3pt]
 & RACE & \scorepair{59.5}{59.6} & \scorepair{60.4}{70.3} & \scorepair{79.5}{89.6} & \scorepair{68.9}{71.2} & \scorepair{48.0}{55.6} & \scorepair{79.9}{84.3} \\
 & Banking77 & \scorepair{92.7}{92.7} & \scorepair{53.4}{66.4} & \scorepair{76.7}{89.2} & \scorepair{27.3}{57.1} & \scorepair{45.4}{56.8} & \scorepair{77.4}{83.2} \\
 & CodeXGLUE-JS & \scorepair{15.8}{15.9} & \scorepair{55.7}{68.7} & \scorepair{79.2}{89.8} & \scorepair{71.3}{71.2} & \scorepair{45.6}{53.0} & \scorepair{78.8}{84.1} \\
\addlinespace[3pt]
 & LEDGAR-Top10 & \scorepair{98.0}{98.2} & \scorepair{53.8}{67.9} & \scorepair{73.1}{89.2} & \scorepair{55.0}{65.6} & \scorepair{46.0}{55.0} & \scorepair{79.4}{83.7} \\
 & SNIPS-7 & \scorepair{98.3}{98.3} & \scorepair{44.8}{63.1} & \scorepair{70.3}{87.6} & \scorepair{61.8}{66.8} & \scorepair{41.8}{52.8} & \scorepair{75.1}{81.7} \\
 & DROP & \scorepair{76.2}{76.1} & \scorepair{59.9}{71.3} & \scorepair{76.1}{88.2} & \scorepair{64.8}{66.5} & \scorepair{46.8}{56.4} & \scorepair{79.8}{83.6} \\
\bottomrule
\end{tabular*}
\par\vspace{6pt}
\begin{tabular*}{\linewidth}{@{\extracolsep{\fill}}ll*{5}{c}@{}}
\toprule
\textbf{Role} & \textbf{Domain} & \textbf{SciQ} & \textbf{WinoGrande} & \textbf{ANLI-R1} & \textbf{SocialIQA} & \textbf{LogiQA} \\
\midrule
Held-out & MedMCQA & \scorepair{95.7}{97.5} & \scorepair{72.0}{78.5} & \scorepair{53.1}{62.8} & \scorepair{59.3}{60.3} & \scorepair{35.5}{44.4} \\
 & MASSIVE-en & \scorepair{96.0}{96.7} & \scorepair{74.2}{78.9} & \scorepair{44.7}{50.6} & \scorepair{50.0}{52.1} & \scorepair{34.9}{43.2} \\
 & WikiSQL & \scorepair{96.9}{97.7} & \scorepair{73.0}{80.9} & \scorepair{51.9}{56.9} & \scorepair{53.2}{56.4} & \scorepair{34.1}{36.7} \\
\midrule
Source & HateXplain & \scorepair{77.6}{95.0} & \scorepair{65.6}{77.4} & \scorepair{40.5}{55.5} & \scorepair{40.5}{48.3} & \scorepair{31.6}{39.9} \\
 & SemEval-2010 T8 & \scorepair{80.9}{95.0} & \scorepair{69.4}{80.0} & \scorepair{50.6}{58.2} & \scorepair{45.9}{52.4} & \scorepair{33.8}{35.6} \\
 & SCOTUS & \scorepair{85.0}{92.6} & \scorepair{73.6}{79.2} & \scorepair{40.4}{45.9} & \scorepair{47.6}{50.7} & \scorepair{31.8}{33.3} \\
\addlinespace[3pt]
 & RACE & \scorepair{96.5}{97.6} & \scorepair{69.1}{78.5} & \scorepair{50.8}{63.8} & \scorepair{57.5}{61.0} & \scorepair{48.2}{48.8} \\
 & Banking77 & \scorepair{95.0}{96.3} & \scorepair{72.6}{80.6} & \scorepair{35.0}{48.6} & \scorepair{47.2}{48.4} & \scorepair{33.2}{36.4} \\
 & CodeXGLUE-JS & \scorepair{96.5}{97.2} & \scorepair{70.7}{79.6} & \scorepair{52.4}{60.7} & \scorepair{54.7}{58.2} & \scorepair{39.3}{46.9} \\
\addlinespace[3pt]
 & LEDGAR-Top10 & \scorepair{88.1}{96.0} & \scorepair{70.8}{78.4} & \scorepair{48.5}{52.9} & \scorepair{49.0}{53.5} & \scorepair{33.6}{34.4} \\
 & SNIPS-7 & \scorepair{87.3}{94.4} & \scorepair{68.7}{78.4} & \scorepair{37.1}{46.9} & \scorepair{38.0}{42.5} & \scorepair{31.0}{36.9} \\
 & DROP & \scorepair{97.3}{97.4} & \scorepair{75.1}{81.2} & \scorepair{47.0}{62.7} & \scorepair{54.9}{58.7} & \scorepair{44.9}{45.3} \\
\bottomrule
\end{tabular*}
\par\vspace{6pt}
\begin{tabular*}{\linewidth}{@{\extracolsep{\fill}}ll*{5}{c}@{}}
\toprule
\textbf{Role} & \textbf{Domain} & \textbf{WiC} & \textbf{COPA} & \textbf{MATH} & \textbf{HumanEval} & \textbf{MBPP} \\
\midrule
Held-out & MedMCQA & \scorepair{55.8}{67.9} & \scorepair{90}{95} & \scorepair{40.2}{44.4} & \scorepair{56.1}{52.4} & \scorepair{62.8}{57.8} \\
 & MASSIVE-en & \scorepair{64.9}{52.7} & \scorepair{91}{95} & \scorepair{43.4}{45.3} & \scorepair{46.3}{50.0} & \scorepair{42.0}{48.6} \\
 & WikiSQL & \scorepair{60.5}{62.9} & \scorepair{91}{94} & \scorepair{39.7}{41.0} & \scorepair{52.4}{43.3} & \scorepair{59.4}{56.4} \\
\midrule
Source & HateXplain & \scorepair{50.0}{56.7} & \scorepair{75}{93} & \scorepair{0.0}{40.4} & \scorepair{0.0}{33.5} & \scorepair{0.0}{45.8} \\
 & SemEval-2010 T8 & \scorepair{49.7}{58.5} & \scorepair{85}{93} & \scorepair{0.0}{41.6} & \scorepair{0.0}{45.1} & \scorepair{6.4}{52.4} \\
 & SCOTUS & \scorepair{50.3}{62.4} & \scorepair{89}{93} & \scorepair{0.0}{43.4} & \scorepair{22.6}{31.7} & \scorepair{36.0}{27.4} \\
\addlinespace[3pt]
 & RACE & \scorepair{51.6}{68.7} & \scorepair{84}{93} & \scorepair{34.7}{43.3} & \scorepair{54.3}{46.3} & \scorepair{62.4}{62.6} \\
 & Banking77 & \scorepair{56.6}{59.1} & \scorepair{90}{94} & \scorepair{15.0}{40.1} & \scorepair{43.3}{44.5} & \scorepair{47.6}{42.0} \\
 & CodeXGLUE-JS & \scorepair{56.0}{54.5} & \scorepair{89}{93} & \scorepair{22.0}{43.7} & \scorepair{53.7}{50.0} & \scorepair{60.4}{58.4} \\
\addlinespace[3pt]
 & LEDGAR-Top10 & \scorepair{50.0}{63.6} & \scorepair{88}{94} & \scorepair{3.6}{38.9} & \scorepair{22.0}{36.6} & \scorepair{28.8}{42.8} \\
 & SNIPS-7 & \scorepair{50.3}{54.5} & \scorepair{85}{94} & \scorepair{0.0}{35.5} & \scorepair{1.8}{31.7} & \scorepair{0.0}{31.4} \\
 & DROP & \scorepair{65.8}{66.1} & \scorepair{90}{94} & \scorepair{33.4}{38.9} & \scorepair{51.8}{46.3} & \scorepair{58.8}{55.2} \\
\bottomrule
\end{tabular*}
\end{center}
\endgroup

\clearpage
\subsection{Split B: Llama-3.1-8B}
\begingroup
\newcommand{\scorepair}[2]{\mbox{#1\(\scriptstyle\rightarrow\)#2}}
\begin{center}
\captionsetup{type=table,font=small,skip=5pt}
\caption{Llama-3.1-8B results on Split B.}
\label{tab:split-llama-B}
\fontsize{8}{9}\selectfont
\setlength{\tabcolsep}{1pt}
\renewcommand{\arraystretch}{1.0}
\begin{tabular*}{\linewidth}{@{\extracolsep{\fill}}ll*{6}{c}@{}}
\toprule
\textbf{Role} & \textbf{Domain} & \textbf{Domain score} & \textbf{ARC-C} & \textbf{HellaSwag} & \textbf{MMLU} & \textbf{OpenBookQA} & \textbf{PIQA} \\
\midrule
Held-out & MedMCQA & \scorepair{40.0}{39.4} & \scorepair{56.5}{61.2} & \scorepair{79.8}{86.6} & \scorepair{45.3}{38.1} & \scorepair{49.0}{54.6} & \scorepair{81.5}{82.8} \\
 & MASSIVE-en & \scorepair{84.2}{84.0} & \scorepair{46.7}{55.8} & \scorepair{75.0}{83.7} & \scorepair{41.1}{40.8} & \scorepair{42.8}{52.4} & \scorepair{77.4}{81.3} \\
 & WikiSQL & \scorepair{85.6}{84.9} & \scorepair{49.5}{55.9} & \scorepair{76.6}{85.4} & \scorepair{52.0}{50.6} & \scorepair{42.4}{52.0} & \scorepair{79.4}{81.8} \\
\midrule
Source & HateXplain & \scorepair{65.6}{65.8} & \scorepair{35.5}{54.5} & \scorepair{51.6}{82.8} & \scorepair{36.8}{51.9} & \scorepair{39.0}{50.4} & \scorepair{67.5}{81.7} \\
 & SemEval-2010 T8 & \scorepair{89.0}{89.0} & \scorepair{42.7}{57.6} & \scorepair{73.9}{85.4} & \scorepair{51.6}{50.7} & \scorepair{43.8}{50.6} & \scorepair{78.1}{82.4} \\
 & SCOTUS & \scorepair{68.5}{68.6} & \scorepair{47.7}{59.9} & \scorepair{78.1}{85.9} & \scorepair{35.4}{46.2} & \scorepair{45.6}{54.0} & \scorepair{79.4}{83.4} \\
\addlinespace[3pt]
 & RACE & \scorepair{59.6}{59.4} & \scorepair{58.8}{64.5} & \scorepair{78.9}{86.1} & \scorepair{47.6}{50.8} & \scorepair{47.0}{54.4} & \scorepair{81.5}{83.4} \\
 & Banking77 & \scorepair{92.9}{92.9} & \scorepair{49.2}{56.8} & \scorepair{77.0}{84.0} & \scorepair{35.0}{38.0} & \scorepair{47.0}{52.2} & \scorepair{78.0}{81.2} \\
 & CodeXGLUE-JS & \scorepair{15.6}{15.6} & \scorepair{53.4}{59.6} & \scorepair{78.8}{86.6} & \scorepair{62.7}{62.2} & \scorepair{46.0}{54.0} & \scorepair{80.6}{82.8} \\
\addlinespace[3pt]
 & LEDGAR-Top10 & \scorepair{98.0}{98.0} & \scorepair{40.1}{54.9} & \scorepair{62.6}{83.1} & \scorepair{28.1}{31.5} & \scorepair{42.8}{51.6} & \scorepair{74.0}{81.5} \\
 & SNIPS-7 & \scorepair{98.7}{98.7} & \scorepair{38.1}{53.0} & \scorepair{68.3}{81.5} & \scorepair{23.5}{28.3} & \scorepair{41.0}{51.4} & \scorepair{75.9}{81.0} \\
 & DROP & \scorepair{65.4}{65.3} & \scorepair{55.2}{61.9} & \scorepair{76.2}{85.0} & \scorepair{33.2}{39.5} & \scorepair{47.2}{56.4} & \scorepair{79.8}{82.6} \\
\bottomrule
\end{tabular*}
\par\vspace{6pt}
\begin{tabular*}{\linewidth}{@{\extracolsep{\fill}}ll*{5}{c}@{}}
\toprule
\textbf{Role} & \textbf{Domain} & \textbf{SciQ} & \textbf{WinoGrande} & \textbf{ANLI-R1} & \textbf{SocialIQA} & \textbf{LogiQA} \\
\midrule
Held-out & MedMCQA & \scorepair{95.7}{96.9} & \scorepair{73.5}{78.5} & \scorepair{34.4}{37.7} & \scorepair{52.5}{53.8} & \scorepair{29.3}{29.3} \\
 & MASSIVE-en & \scorepair{84.4}{89.5} & \scorepair{74.0}{78.0} & \scorepair{32.3}{46.4} & \scorepair{42.2}{43.8} & \scorepair{30.7}{30.1} \\
 & WikiSQL & \scorepair{96.0}{97.0} & \scorepair{73.2}{78.9} & \scorepair{32.6}{36.4} & \scorepair{48.5}{51.7} & \scorepair{30.9}{30.6} \\
\midrule
Source & HateXplain & \scorepair{64.4}{90.5} & \scorepair{66.7}{76.6} & \scorepair{33.3}{41.5} & \scorepair{36.6}{41.3} & \scorepair{26.3}{28.3} \\
 & SemEval-2010 T8 & \scorepair{72.0}{94.9} & \scorepair{71.6}{78.2} & \scorepair{34.9}{46.6} & \scorepair{45.6}{50.0} & \scorepair{29.5}{30.4} \\
 & SCOTUS & \scorepair{80.1}{92.9} & \scorepair{72.2}{77.4} & \scorepair{34.2}{44.8} & \scorepair{47.1}{49.6} & \scorepair{32.0}{30.6} \\
\addlinespace[3pt]
 & RACE & \scorepair{96.3}{97.7} & \scorepair{71.9}{79.3} & \scorepair{35.0}{47.2} & \scorepair{52.0}{55.6} & \scorepair{28.0}{26.9} \\
 & Banking77 & \scorepair{95.5}{95.0} & \scorepair{73.6}{78.5} & \scorepair{37.3}{47.8} & \scorepair{43.9}{47.3} & \scorepair{29.6}{28.7} \\
 & CodeXGLUE-JS & \scorepair{95.6}{97.4} & \scorepair{73.2}{77.8} & \scorepair{36.8}{43.6} & \scorepair{49.6}{52.8} & \scorepair{31.6}{32.1} \\
\addlinespace[3pt]
 & LEDGAR-Top10 & \scorepair{69.0}{93.3} & \scorepair{71.0}{78.2} & \scorepair{36.2}{45.0} & \scorepair{40.7}{45.5} & \scorepair{31.2}{29.5} \\
 & SNIPS-7 & \scorepair{76.7}{90.5} & \scorepair{70.7}{76.9} & \scorepair{33.2}{40.9} & \scorepair{40.3}{44.7} & \scorepair{29.3}{28.7} \\
 & DROP & \scorepair{96.3}{97.2} & \scorepair{74.5}{80.1} & \scorepair{32.1}{43.1} & \scorepair{51.5}{54.0} & \scorepair{27.5}{28.0} \\
\bottomrule
\end{tabular*}
\par\vspace{6pt}
\begin{tabular*}{\linewidth}{@{\extracolsep{\fill}}ll*{5}{c}@{}}
\toprule
\textbf{Role} & \textbf{Domain} & \textbf{WiC} & \textbf{COPA} & \textbf{MATH} & \textbf{HumanEval} & \textbf{MBPP} \\
\midrule
Held-out & MedMCQA & \scorepair{59.9}{58.6} & \scorepair{90}{95} & \scorepair{12.2}{14.7} & \scorepair{34.8}{29.9} & \scorepair{43.2}{35.0} \\
 & MASSIVE-en & \scorepair{50.5}{50.0} & \scorepair{90}{92} & \scorepair{0.0}{9.8} & \scorepair{0.6}{20.1} & \scorepair{0.0}{16.8} \\
 & WikiSQL & \scorepair{50.9}{50.0} & \scorepair{87}{94} & \scorepair{6.1}{10.1} & \scorepair{20.1}{19.5} & \scorepair{28.6}{24.2} \\
\midrule
Source & HateXplain & \scorepair{50.0}{50.2} & \scorepair{71}{93} & \scorepair{0.0}{2.6} & \scorepair{0.0}{14.6} & \scorepair{0.0}{5.0} \\
 & SemEval-2010 T8 & \scorepair{50.0}{50.9} & \scorepair{89}{93} & \scorepair{0.0}{6.8} & \scorepair{0.6}{26.8} & \scorepair{0.0}{29.0} \\
 & SCOTUS & \scorepair{52.2}{53.3} & \scorepair{94}{94} & \scorepair{0.0}{10.6} & \scorepair{28.0}{25.6} & \scorepair{35.2}{26.8} \\
\addlinespace[3pt]
 & RACE & \scorepair{50.2}{62.7} & \scorepair{91}{95} & \scorepair{3.3}{8.1} & \scorepair{34.1}{35.4} & \scorepair{41.6}{35.6} \\
 & Banking77 & \scorepair{49.8}{55.6} & \scorepair{91}{92} & \scorepair{0.0}{10.5} & \scorepair{25.6}{23.8} & \scorepair{29.0}{29.0} \\
 & CodeXGLUE-JS & \scorepair{58.3}{50.5} & \scorepair{93}{95} & \scorepair{5.9}{14.8} & \scorepair{33.5}{29.9} & \scorepair{42.2}{35.2} \\
\addlinespace[3pt]
 & LEDGAR-Top10 & \scorepair{50.0}{50.9} & \scorepair{86}{93} & \scorepair{0.0}{6.7} & \scorepair{0.0}{12.8} & \scorepair{0.0}{11.8} \\
 & SNIPS-7 & \scorepair{49.8}{52.0} & \scorepair{88}{93} & \scorepair{0.0}{4.2} & \scorepair{12.2}{15.9} & \scorepair{1.2}{12.4} \\
 & DROP & \scorepair{53.1}{58.2} & \scorepair{92}{94} & \scorepair{0.0}{3.2} & \scorepair{13.4}{24.4} & \scorepair{25.4}{34.0} \\
\bottomrule
\end{tabular*}
\end{center}
\endgroup

\clearpage
\subsection{Split C: Qwen2.5-7B}
\begingroup
\newcommand{\scorepair}[2]{\mbox{#1\(\scriptstyle\rightarrow\)#2}}
\begin{center}
\captionsetup{type=table,font=small,skip=5pt}
\caption{Qwen2.5-7B results on Split C.}
\label{tab:split-qwen-C}
\fontsize{8}{9}\selectfont
\setlength{\tabcolsep}{1pt}
\renewcommand{\arraystretch}{1.0}
\begin{tabular*}{\linewidth}{@{\extracolsep{\fill}}ll*{6}{c}@{}}
\toprule
\textbf{Role} & \textbf{Domain} & \textbf{Domain score} & \textbf{ARC-C} & \textbf{HellaSwag} & \textbf{MMLU} & \textbf{OpenBookQA} & \textbf{PIQA} \\
\midrule
Held-out & RACE & \scorepair{59.5}{64.0} & \scorepair{60.4}{70.4} & \scorepair{79.5}{90.2} & \scorepair{68.9}{70.6} & \scorepair{48.0}{57.4} & \scorepair{79.9}{83.9} \\
 & Banking77 & \scorepair{92.7}{92.7} & \scorepair{53.4}{64.2} & \scorepair{76.7}{89.4} & \scorepair{27.3}{34.6} & \scorepair{45.4}{54.2} & \scorepair{77.4}{82.5} \\
 & CodeXGLUE-JS & \scorepair{15.8}{15.7} & \scorepair{55.7}{68.1} & \scorepair{79.2}{90.4} & \scorepair{71.3}{71.6} & \scorepair{45.6}{52.2} & \scorepair{78.8}{84.3} \\
\midrule
Source & HateXplain & \scorepair{68.3}{68.6} & \scorepair{41.1}{64.9} & \scorepair{63.0}{89.6} & \scorepair{67.4}{69.1} & \scorepair{43.2}{51.4} & \scorepair{73.3}{83.5} \\
 & SemEval-2010 T8 & \scorepair{87.5}{87.5} & \scorepair{51.4}{68.0} & \scorepair{72.9}{89.5} & \scorepair{63.3}{65.1} & \scorepair{46.6}{56.0} & \scorepair{77.3}{84.4} \\
 & SCOTUS & \scorepair{68.3}{68.4} & \scorepair{45.7}{65.7} & \scorepair{77.7}{89.9} & \scorepair{25.6}{62.5} & \scorepair{46.8}{54.0} & \scorepair{80.2}{83.7} \\
\addlinespace[3pt]
 & MedMCQA & \scorepair{38.4}{38.1} & \scorepair{62.5}{73.5} & \scorepair{79.3}{90.4} & \scorepair{69.5}{70.3} & \scorepair{50.2}{56.8} & \scorepair{81.4}{84.7} \\
 & MASSIVE-en & \scorepair{85.7}{85.7} & \scorepair{56.7}{70.5} & \scorepair{78.1}{90.3} & \scorepair{68.9}{69.3} & \scorepair{47.8}{57.2} & \scorepair{79.7}{85.3} \\
 & WikiSQL & \scorepair{86.5}{86.3} & \scorepair{55.9}{66.4} & \scorepair{77.3}{90.6} & \scorepair{66.6}{67.5} & \scorepair{44.4}{50.8} & \scorepair{79.4}{83.8} \\
\addlinespace[3pt]
 & LEDGAR-Top10 & \scorepair{98.0}{98.2} & \scorepair{53.8}{68.1} & \scorepair{73.1}{89.3} & \scorepair{55.0}{62.1} & \scorepair{46.0}{53.6} & \scorepair{79.4}{85.1} \\
 & SNIPS-7 & \scorepair{98.3}{98.3} & \scorepair{44.8}{64.7} & \scorepair{70.3}{89.5} & \scorepair{61.8}{66.0} & \scorepair{41.8}{52.4} & \scorepair{75.1}{83.2} \\
 & DROP & \scorepair{76.2}{76.1} & \scorepair{59.9}{70.9} & \scorepair{76.1}{89.9} & \scorepair{64.8}{66.4} & \scorepair{46.8}{55.6} & \scorepair{79.8}{85.1} \\
\bottomrule
\end{tabular*}
\par\vspace{6pt}
\begin{tabular*}{\linewidth}{@{\extracolsep{\fill}}ll*{5}{c}@{}}
\toprule
\textbf{Role} & \textbf{Domain} & \textbf{SciQ} & \textbf{WinoGrande} & \textbf{ANLI-R1} & \textbf{SocialIQA} & \textbf{LogiQA} \\
\midrule
Held-out & RACE & \scorepair{96.5}{97.8} & \scorepair{69.1}{77.7} & \scorepair{50.8}{63.3} & \scorepair{57.5}{60.8} & \scorepair{48.2}{46.4} \\
 & Banking77 & \scorepair{95.0}{95.2} & \scorepair{72.6}{79.6} & \scorepair{35.0}{45.5} & \scorepair{47.2}{49.0} & \scorepair{33.2}{37.3} \\
 & CodeXGLUE-JS & \scorepair{96.5}{97.3} & \scorepair{70.7}{79.2} & \scorepair{52.4}{50.3} & \scorepair{54.7}{58.0} & \scorepair{39.3}{43.2} \\
\midrule
Source & HateXplain & \scorepair{77.6}{94.3} & \scorepair{65.6}{78.7} & \scorepair{40.5}{56.4} & \scorepair{40.5}{48.9} & \scorepair{31.6}{39.0} \\
 & SemEval-2010 T8 & \scorepair{80.9}{95.0} & \scorepair{69.4}{79.7} & \scorepair{50.6}{56.8} & \scorepair{45.9}{53.2} & \scorepair{33.8}{34.7} \\
 & SCOTUS & \scorepair{85.0}{93.2} & \scorepair{73.6}{80.4} & \scorepair{40.4}{44.6} & \scorepair{47.6}{50.8} & \scorepair{31.8}{32.6} \\
\addlinespace[3pt]
 & MedMCQA & \scorepair{95.7}{97.6} & \scorepair{72.0}{79.0} & \scorepair{53.1}{62.0} & \scorepair{59.3}{61.6} & \scorepair{35.5}{39.9} \\
 & MASSIVE-en & \scorepair{96.0}{97.2} & \scorepair{74.2}{80.0} & \scorepair{44.7}{52.3} & \scorepair{50.0}{54.4} & \scorepair{34.9}{40.1} \\
 & WikiSQL & \scorepair{96.9}{97.6} & \scorepair{73.0}{81.4} & \scorepair{51.9}{61.9} & \scorepair{53.2}{57.8} & \scorepair{34.1}{40.2} \\
\addlinespace[3pt]
 & LEDGAR-Top10 & \scorepair{88.1}{96.1} & \scorepair{70.8}{78.3} & \scorepair{48.5}{54.4} & \scorepair{49.0}{53.9} & \scorepair{33.6}{33.5} \\
 & SNIPS-7 & \scorepair{87.3}{94.6} & \scorepair{68.7}{79.2} & \scorepair{37.1}{46.9} & \scorepair{38.0}{44.7} & \scorepair{31.0}{38.4} \\
 & DROP & \scorepair{97.3}{97.6} & \scorepair{75.1}{82.1} & \scorepair{47.0}{64.5} & \scorepair{54.9}{59.9} & \scorepair{44.9}{45.2} \\
\bottomrule
\end{tabular*}
\par\vspace{6pt}
\begin{tabular*}{\linewidth}{@{\extracolsep{\fill}}ll*{5}{c}@{}}
\toprule
\textbf{Role} & \textbf{Domain} & \textbf{WiC} & \textbf{COPA} & \textbf{MATH} & \textbf{HumanEval} & \textbf{MBPP} \\
\midrule
Held-out & RACE & \scorepair{51.6}{64.4} & \scorepair{84}{92} & \scorepair{34.7}{44.2} & \scorepair{54.3}{50.6} & \scorepair{62.4}{60.4} \\
 & Banking77 & \scorepair{56.6}{56.9} & \scorepair{90}{92} & \scorepair{15.0}{38.9} & \scorepair{43.3}{43.3} & \scorepair{47.6}{45.2} \\
 & CodeXGLUE-JS & \scorepair{56.0}{52.8} & \scorepair{89}{92} & \scorepair{22.0}{45.4} & \scorepair{53.7}{50.6} & \scorepair{60.4}{57.8} \\
\midrule
Source & HateXplain & \scorepair{50.0}{57.4} & \scorepair{75}{93} & \scorepair{0.0}{41.0} & \scorepair{0.0}{34.1} & \scorepair{0.0}{46.6} \\
 & SemEval-2010 T8 & \scorepair{49.7}{59.1} & \scorepair{85}{95} & \scorepair{0.0}{41.0} & \scorepair{0.0}{47.6} & \scorepair{6.4}{52.2} \\
 & SCOTUS & \scorepair{50.3}{62.9} & \scorepair{89}{94} & \scorepair{0.0}{43.1} & \scorepair{22.6}{29.9} & \scorepair{36.0}{29.2} \\
\addlinespace[3pt]
 & MedMCQA & \scorepair{55.8}{67.4} & \scorepair{90}{94} & \scorepair{40.2}{44.6} & \scorepair{56.1}{49.4} & \scorepair{62.8}{61.0} \\
 & MASSIVE-en & \scorepair{64.9}{60.7} & \scorepair{91}{96} & \scorepair{43.4}{45.6} & \scorepair{46.3}{43.9} & \scorepair{42.0}{46.4} \\
 & WikiSQL & \scorepair{60.5}{53.1} & \scorepair{91}{94} & \scorepair{39.7}{40.8} & \scorepair{52.4}{47.0} & \scorepair{59.4}{57.4} \\
\addlinespace[3pt]
 & LEDGAR-Top10 & \scorepair{50.0}{63.3} & \scorepair{88}{93} & \scorepair{3.6}{38.3} & \scorepair{22.0}{39.6} & \scorepair{28.8}{44.2} \\
 & SNIPS-7 & \scorepair{50.3}{55.3} & \scorepair{85}{95} & \scorepair{0.0}{35.2} & \scorepair{1.8}{36.0} & \scorepair{0.0}{30.4} \\
 & DROP & \scorepair{65.8}{65.0} & \scorepair{90}{97} & \scorepair{33.4}{39.3} & \scorepair{51.8}{48.2} & \scorepair{58.8}{58.0} \\
\bottomrule
\end{tabular*}
\end{center}
\endgroup

\clearpage
\subsection{Split C: Llama-3.1-8B}
\begingroup
\newcommand{\scorepair}[2]{\mbox{#1\(\scriptstyle\rightarrow\)#2}}
\begin{center}
\captionsetup{type=table,font=small,skip=5pt}
\caption{Llama-3.1-8B results on Split C.}
\label{tab:split-llama-C}
\fontsize{8}{9}\selectfont
\setlength{\tabcolsep}{1pt}
\renewcommand{\arraystretch}{1.0}
\begin{tabular*}{\linewidth}{@{\extracolsep{\fill}}ll*{6}{c}@{}}
\toprule
\textbf{Role} & \textbf{Domain} & \textbf{Domain score} & \textbf{ARC-C} & \textbf{HellaSwag} & \textbf{MMLU} & \textbf{OpenBookQA} & \textbf{PIQA} \\
\midrule
Held-out & RACE & \scorepair{59.6}{59.6} & \scorepair{58.8}{60.1} & \scorepair{78.9}{84.4} & \scorepair{47.6}{48.3} & \scorepair{47.0}{51.4} & \scorepair{81.5}{81.4} \\
 & Banking77 & \scorepair{92.9}{92.9} & \scorepair{49.2}{54.9} & \scorepair{77.0}{84.0} & \scorepair{35.0}{37.3} & \scorepair{47.0}{53.0} & \scorepair{78.0}{81.0} \\
 & CodeXGLUE-JS & \scorepair{15.6}{15.6} & \scorepair{53.4}{59.4} & \scorepair{78.8}{86.2} & \scorepair{62.7}{62.3} & \scorepair{46.0}{52.0} & \scorepair{80.6}{81.8} \\
\midrule
Source & HateXplain & \scorepair{65.6}{65.8} & \scorepair{35.5}{54.5} & \scorepair{51.6}{83.5} & \scorepair{36.8}{54.1} & \scorepair{39.0}{52.0} & \scorepair{67.5}{80.7} \\
 & SemEval-2010 T8 & \scorepair{89.0}{89.0} & \scorepair{42.7}{56.7} & \scorepair{73.9}{86.3} & \scorepair{51.6}{53.5} & \scorepair{43.8}{51.2} & \scorepair{78.1}{82.9} \\
 & SCOTUS & \scorepair{68.5}{68.6} & \scorepair{47.7}{59.5} & \scorepair{78.1}{86.8} & \scorepair{35.4}{50.1} & \scorepair{45.6}{51.6} & \scorepair{79.4}{82.5} \\
\addlinespace[3pt]
 & MedMCQA & \scorepair{40.0}{39.9} & \scorepair{56.5}{61.9} & \scorepair{79.8}{86.6} & \scorepair{45.3}{54.2} & \scorepair{49.0}{53.4} & \scorepair{81.5}{82.6} \\
 & MASSIVE-en & \scorepair{84.2}{84.2} & \scorepair{46.7}{57.4} & \scorepair{75.0}{85.4} & \scorepair{41.1}{45.7} & \scorepair{42.8}{51.6} & \scorepair{77.4}{81.0} \\
 & WikiSQL & \scorepair{85.6}{85.5} & \scorepair{49.5}{55.8} & \scorepair{76.6}{86.3} & \scorepair{52.0}{54.9} & \scorepair{42.4}{51.0} & \scorepair{79.4}{81.5} \\
\addlinespace[3pt]
 & LEDGAR-Top10 & \scorepair{98.0}{98.0} & \scorepair{40.1}{55.0} & \scorepair{62.6}{83.4} & \scorepair{28.1}{32.8} & \scorepair{42.8}{51.8} & \scorepair{74.0}{81.1} \\
 & SNIPS-7 & \scorepair{98.7}{98.7} & \scorepair{38.1}{51.9} & \scorepair{68.3}{82.0} & \scorepair{23.5}{30.0} & \scorepair{41.0}{50.0} & \scorepair{75.9}{79.6} \\
 & DROP & \scorepair{65.4}{65.3} & \scorepair{55.2}{60.0} & \scorepair{76.2}{84.8} & \scorepair{33.2}{40.2} & \scorepair{47.2}{54.2} & \scorepair{79.8}{82.4} \\
\bottomrule
\end{tabular*}
\par\vspace{6pt}
\begin{tabular*}{\linewidth}{@{\extracolsep{\fill}}ll*{5}{c}@{}}
\toprule
\textbf{Role} & \textbf{Domain} & \textbf{SciQ} & \textbf{WinoGrande} & \textbf{ANLI-R1} & \textbf{SocialIQA} & \textbf{LogiQA} \\
\midrule
Held-out & RACE & \scorepair{96.3}{97.5} & \scorepair{71.9}{78.2} & \scorepair{35.0}{34.7} & \scorepair{52.0}{55.0} & \scorepair{28.0}{28.7} \\
 & Banking77 & \scorepair{95.5}{95.8} & \scorepair{73.6}{79.2} & \scorepair{37.3}{39.1} & \scorepair{43.9}{45.9} & \scorepair{29.6}{29.2} \\
 & CodeXGLUE-JS & \scorepair{95.6}{97.2} & \scorepair{73.2}{78.1} & \scorepair{36.8}{43.9} & \scorepair{49.6}{51.9} & \scorepair{31.6}{32.0} \\
\midrule
Source & HateXplain & \scorepair{64.4}{91.7} & \scorepair{66.7}{76.7} & \scorepair{33.3}{40.8} & \scorepair{36.6}{40.9} & \scorepair{26.3}{29.0} \\
 & SemEval-2010 T8 & \scorepair{72.0}{93.7} & \scorepair{71.6}{78.3} & \scorepair{34.9}{43.2} & \scorepair{45.6}{49.9} & \scorepair{29.5}{30.7} \\
 & SCOTUS & \scorepair{80.1}{93.5} & \scorepair{72.2}{78.0} & \scorepair{34.2}{43.6} & \scorepair{47.1}{50.0} & \scorepair{32.0}{30.4} \\
\addlinespace[3pt]
 & MedMCQA & \scorepair{95.7}{96.7} & \scorepair{73.5}{79.0} & \scorepair{34.4}{41.9} & \scorepair{52.5}{55.2} & \scorepair{29.3}{29.3} \\
 & MASSIVE-en & \scorepair{84.4}{95.0} & \scorepair{74.0}{78.0} & \scorepair{32.3}{48.0} & \scorepair{42.2}{44.1} & \scorepair{30.7}{28.4} \\
 & WikiSQL & \scorepair{96.0}{97.1} & \scorepair{73.2}{79.7} & \scorepair{32.6}{40.0} & \scorepair{48.5}{51.4} & \scorepair{30.9}{30.4} \\
\addlinespace[3pt]
 & LEDGAR-Top10 & \scorepair{69.0}{93.6} & \scorepair{71.0}{78.1} & \scorepair{36.2}{46.0} & \scorepair{40.7}{45.0} & \scorepair{31.2}{29.0} \\
 & SNIPS-7 & \scorepair{76.7}{91.2} & \scorepair{70.7}{76.6} & \scorepair{33.2}{40.3} & \scorepair{40.3}{44.8} & \scorepair{29.3}{29.0} \\
 & DROP & \scorepair{96.3}{97.3} & \scorepair{74.5}{79.8} & \scorepair{32.1}{39.9} & \scorepair{51.5}{53.8} & \scorepair{27.5}{29.5} \\
\bottomrule
\end{tabular*}
\par\vspace{6pt}
\begin{tabular*}{\linewidth}{@{\extracolsep{\fill}}ll*{5}{c}@{}}
\toprule
\textbf{Role} & \textbf{Domain} & \textbf{WiC} & \textbf{COPA} & \textbf{MATH} & \textbf{HumanEval} & \textbf{MBPP} \\
\midrule
Held-out & RACE & \scorepair{50.2}{50.8} & \scorepair{91}{96} & \scorepair{3.3}{7.1} & \scorepair{34.1}{34.1} & \scorepair{41.6}{38.0} \\
 & Banking77 & \scorepair{49.8}{54.2} & \scorepair{91}{93} & \scorepair{0.0}{10.8} & \scorepair{25.6}{25.0} & \scorepair{29.0}{31.6} \\
 & CodeXGLUE-JS & \scorepair{58.3}{58.6} & \scorepair{93}{94} & \scorepair{5.9}{14.2} & \scorepair{33.5}{29.9} & \scorepair{42.2}{36.8} \\
\midrule
Source & HateXplain & \scorepair{50.0}{56.1} & \scorepair{71}{92} & \scorepair{0.0}{1.9} & \scorepair{0.0}{13.4} & \scorepair{0.0}{6.2} \\
 & SemEval-2010 T8 & \scorepair{50.0}{56.7} & \scorepair{89}{91} & \scorepair{0.0}{6.9} & \scorepair{0.6}{25.0} & \scorepair{0.0}{26.2} \\
 & SCOTUS & \scorepair{52.2}{55.2} & \scorepair{94}{94} & \scorepair{0.0}{10.7} & \scorepair{28.0}{28.0} & \scorepair{35.2}{27.4} \\
\addlinespace[3pt]
 & MedMCQA & \scorepair{59.9}{61.6} & \scorepair{90}{94} & \scorepair{12.2}{14.9} & \scorepair{34.8}{30.5} & \scorepair{43.2}{36.0} \\
 & MASSIVE-en & \scorepair{50.5}{53.4} & \scorepair{90}{94} & \scorepair{0.0}{9.6} & \scorepair{0.6}{17.7} & \scorepair{0.0}{14.6} \\
 & WikiSQL & \scorepair{50.9}{50.9} & \scorepair{87}{94} & \scorepair{6.1}{10.0} & \scorepair{20.1}{21.3} & \scorepair{28.6}{26.0} \\
\addlinespace[3pt]
 & LEDGAR-Top10 & \scorepair{50.0}{58.2} & \scorepair{86}{95} & \scorepair{0.0}{6.9} & \scorepair{0.0}{12.8} & \scorepair{0.0}{13.4} \\
 & SNIPS-7 & \scorepair{49.8}{54.1} & \scorepair{88}{93} & \scorepair{0.0}{4.2} & \scorepair{12.2}{17.1} & \scorepair{1.2}{11.0} \\
 & DROP & \scorepair{53.1}{59.4} & \scorepair{92}{94} & \scorepair{0.0}{2.5} & \scorepair{13.4}{23.2} & \scorepair{25.4}{34.4} \\
\bottomrule
\end{tabular*}
\end{center}
\endgroup

\clearpage
\section{Split D Source-domain Results}
\label{app:source-results}

Source models provide functional supervision, \repairman{} parameter targets, and template initialization.

\subsection{Qwen2.5-7B}
\begingroup
\newcommand{\scorepair}[2]{\mbox{#1\(\scriptstyle\rightarrow\)#2}}
\begin{center}
\captionsetup{type=table,font=small,skip=5pt}
\caption{Qwen2.5-7B results on the source domains of Split D.}
\label{tab:source-qwen-D}
\fontsize{8}{9}\selectfont
\setlength{\tabcolsep}{1pt}
\renewcommand{\arraystretch}{1.0}
\begin{tabular*}{\linewidth}{@{\extracolsep{\fill}}l*{6}{c}@{}}
\toprule
\textbf{Domain} & \textbf{Domain score} & \textbf{ARC-C} & \textbf{HellaSwag} & \textbf{MMLU} & \textbf{OpenBookQA} & \textbf{PIQA} \\
\midrule
HateXplain & \scorepair{68.3}{68.6} & \scorepair{41.1}{64.3} & \scorepair{63.0}{89.2} & \scorepair{67.4}{69.0} & \scorepair{43.2}{55.2} & \scorepair{73.3}{83.2} \\
SemEval-2010 T8 & \scorepair{87.5}{87.5} & \scorepair{51.4}{66.8} & \scorepair{72.9}{88.3} & \scorepair{63.3}{62.4} & \scorepair{46.6}{53.4} & \scorepair{77.3}{83.5} \\
SCOTUS & \scorepair{68.3}{68.4} & \scorepair{45.7}{64.1} & \scorepair{77.7}{89.6} & \scorepair{25.6}{63.9} & \scorepair{46.8}{53.4} & \scorepair{80.2}{83.0} \\
\addlinespace[3pt]
MedMCQA & \scorepair{38.4}{38.1} & \scorepair{62.5}{72.9} & \scorepair{79.3}{90.1} & \scorepair{69.5}{70.3} & \scorepair{50.2}{56.2} & \scorepair{81.4}{84.3} \\
MASSIVE-en & \scorepair{85.7}{85.7} & \scorepair{56.7}{70.5} & \scorepair{78.1}{89.6} & \scorepair{68.9}{69.2} & \scorepair{47.8}{57.0} & \scorepair{79.7}{85.3} \\
WikiSQL & \scorepair{86.5}{86.3} & \scorepair{55.9}{66.0} & \scorepair{77.3}{90.0} & \scorepair{66.6}{67.0} & \scorepair{44.4}{51.4} & \scorepair{79.4}{83.8} \\
\addlinespace[3pt]
RACE & \scorepair{59.5}{59.6} & \scorepair{60.4}{70.3} & \scorepair{79.5}{90.1} & \scorepair{68.9}{70.3} & \scorepair{48.0}{55.8} & \scorepair{79.9}{85.0} \\
Banking77 & \scorepair{92.7}{92.7} & \scorepair{53.4}{65.1} & \scorepair{76.7}{89.0} & \scorepair{27.3}{57.2} & \scorepair{45.4}{55.8} & \scorepair{77.4}{83.5} \\
CodeXGLUE-JS & \scorepair{15.8}{15.9} & \scorepair{55.7}{69.6} & \scorepair{79.2}{90.7} & \scorepair{71.3}{71.4} & \scorepair{45.6}{53.4} & \scorepair{78.8}{84.9} \\
\bottomrule
\end{tabular*}
\par\vspace{6pt}
\begin{tabular*}{\linewidth}{@{\extracolsep{\fill}}l*{5}{c}@{}}
\toprule
\textbf{Domain} & \textbf{SciQ} & \textbf{WinoGrande} & \textbf{ANLI-R1} & \textbf{SocialIQA} & \textbf{LogiQA} \\
\midrule
HateXplain & \scorepair{77.6}{94.6} & \scorepair{65.6}{78.1} & \scorepair{40.5}{54.2} & \scorepair{40.5}{47.6} & \scorepair{31.6}{40.4} \\
SemEval-2010 T8 & \scorepair{80.9}{95.5} & \scorepair{69.4}{79.3} & \scorepair{50.6}{58.3} & \scorepair{45.9}{51.9} & \scorepair{33.8}{35.0} \\
SCOTUS & \scorepair{85.0}{90.9} & \scorepair{73.6}{79.6} & \scorepair{40.4}{44.3} & \scorepair{47.6}{49.9} & \scorepair{31.8}{34.6} \\
\addlinespace[3pt]
MedMCQA & \scorepair{95.7}{97.5} & \scorepair{72.0}{79.1} & \scorepair{53.1}{61.9} & \scorepair{59.3}{60.2} & \scorepair{35.5}{43.9} \\
MASSIVE-en & \scorepair{96.0}{97.1} & \scorepair{74.2}{79.6} & \scorepair{44.7}{54.3} & \scorepair{50.0}{52.8} & \scorepair{34.9}{44.1} \\
WikiSQL & \scorepair{96.9}{97.6} & \scorepair{73.0}{81.1} & \scorepair{51.9}{60.6} & \scorepair{53.2}{57.9} & \scorepair{34.1}{38.9} \\
\addlinespace[3pt]
RACE & \scorepair{96.5}{97.6} & \scorepair{69.1}{79.3} & \scorepair{50.8}{65.5} & \scorepair{57.5}{60.9} & \scorepair{48.2}{48.5} \\
Banking77 & \scorepair{95.0}{95.5} & \scorepair{72.6}{80.7} & \scorepair{35.0}{46.7} & \scorepair{47.2}{49.3} & \scorepair{33.2}{37.2} \\
CodeXGLUE-JS & \scorepair{96.5}{97.3} & \scorepair{70.7}{79.8} & \scorepair{52.4}{60.6} & \scorepair{54.7}{57.7} & \scorepair{39.3}{45.8} \\
\bottomrule
\end{tabular*}
\par\vspace{6pt}
\begin{tabular*}{\linewidth}{@{\extracolsep{\fill}}l*{5}{c}@{}}
\toprule
\textbf{Domain} & \textbf{WiC} & \textbf{COPA} & \textbf{MATH} & \textbf{HumanEval} & \textbf{MBPP} \\
\midrule
HateXplain & \scorepair{50.0}{55.8} & \scorepair{75}{93} & \scorepair{0.0}{41.3} & \scorepair{0.0}{32.3} & \scorepair{0.0}{45.4} \\
SemEval-2010 T8 & \scorepair{49.7}{60.2} & \scorepair{85}{94} & \scorepair{0.0}{40.8} & \scorepair{0.0}{45.7} & \scorepair{6.4}{52.2} \\
SCOTUS & \scorepair{50.3}{59.6} & \scorepair{89}{94} & \scorepair{0.0}{43.4} & \scorepair{22.6}{28.7} & \scorepair{36.0}{27.0} \\
\addlinespace[3pt]
MedMCQA & \scorepair{55.8}{68.2} & \scorepair{90}{92} & \scorepair{40.2}{44.4} & \scorepair{56.1}{49.4} & \scorepair{62.8}{59.2} \\
MASSIVE-en & \scorepair{64.9}{61.0} & \scorepair{91}{94} & \scorepair{43.4}{45.6} & \scorepair{46.3}{37.8} & \scorepair{42.0}{46.6} \\
WikiSQL & \scorepair{60.5}{62.9} & \scorepair{91}{92} & \scorepair{39.7}{40.9} & \scorepair{52.4}{51.2} & \scorepair{59.4}{58.8} \\
\addlinespace[3pt]
RACE & \scorepair{51.6}{65.8} & \scorepair{84}{92} & \scorepair{34.7}{43.4} & \scorepair{54.3}{49.4} & \scorepair{62.4}{61.4} \\
Banking77 & \scorepair{56.6}{55.8} & \scorepair{90}{93} & \scorepair{15.0}{39.0} & \scorepair{43.3}{40.9} & \scorepair{47.6}{41.6} \\
CodeXGLUE-JS & \scorepair{56.0}{64.6} & \scorepair{89}{91} & \scorepair{22.0}{45.9} & \scorepair{53.7}{50.0} & \scorepair{60.4}{57.6} \\
\bottomrule
\end{tabular*}
\end{center}
\endgroup
\clearpage
\subsection{Llama-3.1-8B}
\begingroup
\newcommand{\scorepair}[2]{\mbox{#1\(\scriptstyle\rightarrow\)#2}}
\begin{center}
\captionsetup{type=table,font=small,skip=5pt}
\caption{Llama-3.1-8B results on the source domains of Split D.}
\label{tab:source-llama-D}
\fontsize{8}{9}\selectfont
\setlength{\tabcolsep}{1pt}
\renewcommand{\arraystretch}{1.0}
\begin{tabular*}{\linewidth}{@{\extracolsep{\fill}}l*{6}{c}@{}}
\toprule
\textbf{Domain} & \textbf{Domain score} & \textbf{ARC-C} & \textbf{HellaSwag} & \textbf{MMLU} & \textbf{OpenBookQA} & \textbf{PIQA} \\
\midrule
HateXplain & \scorepair{65.6}{65.8} & \scorepair{35.5}{52.7} & \scorepair{51.6}{80.3} & \scorepair{36.8}{51.2} & \scorepair{39.0}{48.4} & \scorepair{67.5}{80.8} \\
SemEval-2010 T8 & \scorepair{89.0}{89.0} & \scorepair{42.7}{56.7} & \scorepair{73.9}{84.3} & \scorepair{51.6}{48.7} & \scorepair{43.8}{51.6} & \scorepair{78.1}{83.4} \\
SCOTUS & \scorepair{68.5}{68.6} & \scorepair{47.7}{58.9} & \scorepair{78.1}{84.8} & \scorepair{35.4}{46.7} & \scorepair{45.6}{51.0} & \scorepair{79.4}{83.4} \\
\addlinespace[3pt]
MedMCQA & \scorepair{40.0}{39.9} & \scorepair{56.5}{62.7} & \scorepair{79.8}{85.3} & \scorepair{45.3}{51.9} & \scorepair{49.0}{54.0} & \scorepair{81.5}{83.2} \\
MASSIVE-en & \scorepair{84.2}{84.2} & \scorepair{46.7}{56.5} & \scorepair{75.0}{83.1} & \scorepair{41.1}{41.3} & \scorepair{42.8}{51.4} & \scorepair{77.4}{81.7} \\
WikiSQL & \scorepair{85.6}{85.5} & \scorepair{49.5}{55.4} & \scorepair{76.6}{84.2} & \scorepair{52.0}{54.0} & \scorepair{42.4}{50.2} & \scorepair{79.4}{82.3} \\
\addlinespace[3pt]
RACE & \scorepair{59.6}{59.4} & \scorepair{58.8}{64.1} & \scorepair{78.9}{85.1} & \scorepair{47.6}{50.0} & \scorepair{47.0}{54.0} & \scorepair{81.5}{83.4} \\
Banking77 & \scorepair{92.9}{92.9} & \scorepair{49.2}{55.0} & \scorepair{77.0}{82.7} & \scorepair{35.0}{37.9} & \scorepair{47.0}{52.6} & \scorepair{78.0}{80.7} \\
CodeXGLUE-JS & \scorepair{15.6}{15.6} & \scorepair{53.4}{61.2} & \scorepair{78.8}{85.9} & \scorepair{62.7}{61.2} & \scorepair{46.0}{53.0} & \scorepair{80.6}{83.2} \\
\bottomrule
\end{tabular*}
\par\vspace{6pt}
\begin{tabular*}{\linewidth}{@{\extracolsep{\fill}}l*{5}{c}@{}}
\toprule
\textbf{Domain} & \textbf{SciQ} & \textbf{WinoGrande} & \textbf{ANLI-R1} & \textbf{SocialIQA} & \textbf{LogiQA} \\
\midrule
HateXplain & \scorepair{64.4}{90.7} & \scorepair{66.7}{76.8} & \scorepair{33.3}{38.3} & \scorepair{36.6}{40.4} & \scorepair{26.3}{28.7} \\
SemEval-2010 T8 & \scorepair{72.0}{94.3} & \scorepair{71.6}{78.1} & \scorepair{34.9}{46.2} & \scorepair{45.6}{49.2} & \scorepair{29.5}{33.2} \\
SCOTUS & \scorepair{80.1}{92.7} & \scorepair{72.2}{78.5} & \scorepair{34.2}{44.2} & \scorepair{47.1}{48.0} & \scorepair{32.0}{30.1} \\
\addlinespace[3pt]
MedMCQA & \scorepair{95.7}{97.1} & \scorepair{73.5}{79.4} & \scorepair{34.4}{45.3} & \scorepair{52.5}{53.4} & \scorepair{29.3}{27.5} \\
MASSIVE-en & \scorepair{84.4}{94.2} & \scorepair{74.0}{78.4} & \scorepair{32.3}{44.7} & \scorepair{42.2}{44.6} & \scorepair{30.7}{28.9} \\
WikiSQL & \scorepair{96.0}{96.7} & \scorepair{73.2}{79.3} & \scorepair{32.6}{39.9} & \scorepair{48.5}{50.1} & \scorepair{30.9}{31.0} \\
\addlinespace[3pt]
RACE & \scorepair{96.3}{97.5} & \scorepair{71.9}{79.2} & \scorepair{35.0}{43.8} & \scorepair{52.0}{54.6} & \scorepair{28.0}{28.9} \\
Banking77 & \scorepair{95.5}{95.8} & \scorepair{73.6}{78.9} & \scorepair{37.3}{46.1} & \scorepair{43.9}{46.5} & \scorepair{29.6}{29.5} \\
CodeXGLUE-JS & \scorepair{95.6}{97.8} & \scorepair{73.2}{78.9} & \scorepair{36.8}{43.8} & \scorepair{49.6}{50.4} & \scorepair{31.6}{33.0} \\
\bottomrule
\end{tabular*}
\par\vspace{6pt}
\begin{tabular*}{\linewidth}{@{\extracolsep{\fill}}l*{5}{c}@{}}
\toprule
\textbf{Domain} & \textbf{WiC} & \textbf{COPA} & \textbf{MATH} & \textbf{HumanEval} & \textbf{MBPP} \\
\midrule
HateXplain & \scorepair{50.0}{55.5} & \scorepair{71}{92} & \scorepair{0.0}{1.1} & \scorepair{0.0}{14.6} & \scorepair{0.0}{4.4} \\
SemEval-2010 T8 & \scorepair{50.0}{59.7} & \scorepair{89}{93} & \scorepair{0.0}{6.6} & \scorepair{0.6}{22.6} & \scorepair{0.0}{24.4} \\
SCOTUS & \scorepair{52.2}{54.9} & \scorepair{94}{95} & \scorepair{0.0}{8.8} & \scorepair{28.0}{29.3} & \scorepair{35.2}{25.2} \\
\addlinespace[3pt]
MedMCQA & \scorepair{59.9}{60.8} & \scorepair{90}{95} & \scorepair{12.2}{14.4} & \scorepair{34.8}{31.7} & \scorepair{43.2}{35.0} \\
MASSIVE-en & \scorepair{50.5}{56.6} & \scorepair{90}{92} & \scorepair{0.0}{8.7} & \scorepair{0.6}{18.9} & \scorepair{0.0}{14.4} \\
WikiSQL & \scorepair{50.9}{54.7} & \scorepair{87}{90} & \scorepair{6.1}{9.7} & \scorepair{20.1}{22.0} & \scorepair{28.6}{25.0} \\
\addlinespace[3pt]
RACE & \scorepair{50.2}{58.0} & \scorepair{91}{97} & \scorepair{3.3}{6.3} & \scorepair{34.1}{32.3} & \scorepair{41.6}{38.4} \\
Banking77 & \scorepair{49.8}{52.5} & \scorepair{91}{92} & \scorepair{0.0}{9.5} & \scorepair{25.6}{26.8} & \scorepair{29.0}{30.4} \\
CodeXGLUE-JS & \scorepair{58.3}{61.3} & \scorepair{93}{93} & \scorepair{5.9}{15.5} & \scorepair{33.5}{28.7} & \scorepair{42.2}{35.6} \\
\bottomrule
\end{tabular*}
\end{center}
\endgroup

\end{document}